\pdfoutput=1

\documentclass[11pt]{article}
\usepackage{CJKutf8}
\usepackage[]{acl}
\usepackage{graphicx}
\usepackage{times}
\usepackage{booktabs}  
\usepackage{wrapfig}
\usepackage{latexsym}
\usepackage{adjustbox}
\usepackage{multirow}
\usepackage{enumitem}
\usepackage[T1]{fontenc}

\usepackage[utf8]{inputenc}
\usepackage{makecell}
\usepackage[figuresright]{rotating} 
\usepackage{microtype}
\usepackage{todonotes}
\usepackage{inconsolata}
\usepackage{tabularx}
\usepackage{threeparttable}

\title{ESC-Eval: Evaluating Emotion Support Conversations\\ in Large Language Models}

\author{
Haiquan Zhao\thanks{ {} Work done during internship at Shanghai Artificial Intellignece Laboratory}$^{1,2}$\ \hspace{.3em}\
Lingyu Li$^{1}$ \hspace{.3em}\
Shisong Chen$^{2}$ \hspace{.3em}\
Shuqi Kong$^{1}$ \hspace{.3em} \
Jiaan Wang $^{2}$ \hspace{.3em}\\
\textbf{
Kexin Huang$^{1,2}$ \hspace{.3em}\
Tianle Gu$^{1}$ \hspace{.3em}\
Yixu Wang$^{1,2}$ \hspace{.3em}\
Jian Wang$^{1}$ \hspace{.3em}\
Dandan Liang$^{1}$ \hspace{.3em} } \\
\textbf{
Zhixu Li\thanks{ {} Correspondence to: Zhixu Li<zhixuli@ruc.edu.cn> and  Yan Teng <tengyan@pjlab.org.cn>}$^{3,4}$ \
Yan Teng\footnotemark[2]$^{1}$ \hspace{.3em} \
Yanghua Xiao$^{2}$ \hspace{.3em}\
Yingchun Wang$^{1}$ \hspace{.3em}\
}\\
$^{1}$ Shanghai Artificial Intelligence Laboratory \\
$^{2}$ School of Computer Science, Fudan University \\
$^{3}$ School of Information, Renmin University of China \\
$^{4}$ Suzhou Key Laboratory of Artificial Intelligence and Social Governance Technologies, \\ International College (Suzhou Research Institute), Renmin
University of China \\
}

\begin{document}
\maketitle
\begin{abstract}
Emotion Support Conversation (ESC) is a crucial application, which aims to reduce human stress, offer emotional guidance, and ultimately enhance human mental and physical well-being.
With the advancement of Large Language Models (LLMs), many researchers have employed LLMs as the ESC models.
However, the evaluation of these LLM-based ESCs remains uncertain.
Inspired by the awesome development of role-playing agents, we propose an ESC Evaluation framework (\emph{i.e.}, ESC-Eval), which uses a role-playing agent to interact with ESC models, followed by a manual evaluation of the interactive dialogues.
In detail, we first re-organize 2,801 role-playing cards from seven existing datasets to define the roles of the role-playing agent.
Second, we train a specific role-playing model - ESC-Role to mimic the behavior of a real person experiencing distress.
Third, through ESC-Role and organized role cards, we systematically conduct experiments using 14 LLMs as the ESC models, including general AI-assistant LLMs (\emph{e.g.}, ChatGPT) and ESC-oriented LLMs (\emph{e.g.}, ExTES-Llama). 
We conduct comprehensive human annotations on interactive multi-turn dialogues of different ESC models. The results show that ESC-oriented LLMs exhibit superior ESC abilities compared to general AI-assistant LLMs, but there is still a gap behind human performance.
Moreover, to automate the evaluation of future ESC models, we developed ESC-RANK, 
which trained on the annotated data, achieving a scoring performance surpassing 35 points of GPT-4. Our data and code are available at \href{https://github.com/AIFlames/Esc-Eval}{https://github.com/AIFlames/Esc-Eval}.
\end{abstract}

\section{Introduction}
\begin{figure*}
\centering
\includegraphics[width=\textwidth]{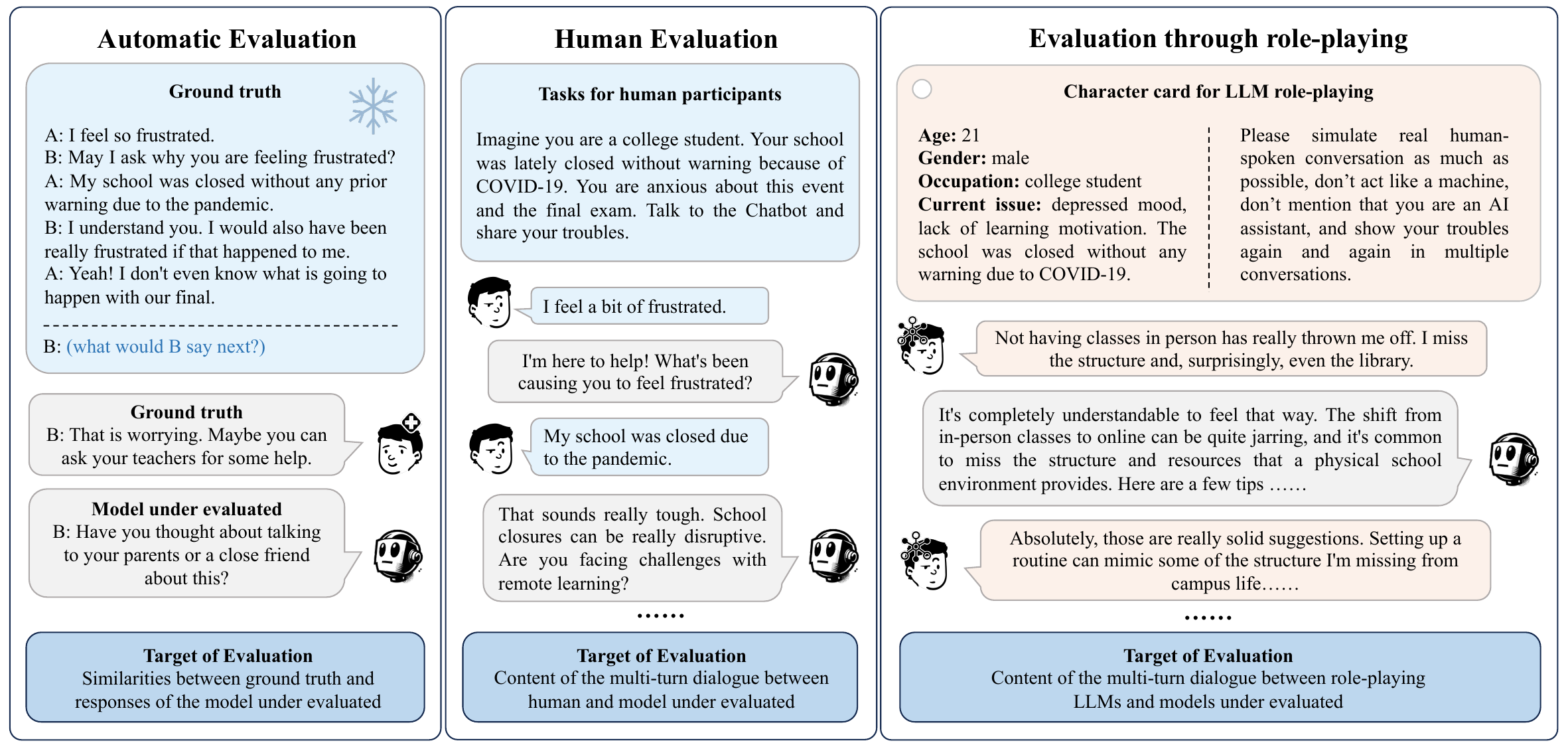}
\caption{Difference between our proposed evaluation framework and others. }
\label{F_intro}
\end{figure*}
With the rapid development of Large Language Models (LLMs), an increasing number of individuals are engaging in conversations with LLMs (\emph{e.g.}, ChatGPT~\cite{achiam2023gpt}).
Among various conversational applications, Emotional Support Conversation~\cite{liu2021towards} (ESC) stands out as a particularly promising field, where people can freely share their personal experiences or concerns, receiving emotional support and practical advice. This interaction helps alleviate human pressures~\cite{langford1997social,burleson2003emotional}, thereby improving overall well-being.
Recently, numerous LLM-based ESC models have received wide research attention~\cite{zheng2023building,qiu2023smile,liu2023chatcounselor}. However, effective and comprehensive evaluation of these chatbots remains challenging.

Current ESC evaluation~\cite{liu2021towards,zheng2023building} generally uses text-based statistical metrics or manual evaluations.
(1) When using text-based statistical metrics, researchers provide the dialogue history to the ESC models and then use the models to generate the corresponding responses (c.f., left panel in Figure~\ref{F_intro}).
Based on the generated responses, text-based statistical metrics (such as ROUGE~\cite{lin2004rouge} and BLEU~\cite{papineni2002bleu}) assess whether the responses resemble ground truth.
However, these metrics heavily rely on ground truth responses, which lack objectivity~\cite{novikova2017we} due to the complex nature of ESC. 
Furthermore, since the conversation history from ground truth is provided to the model under evaluation, text-based statistical metrics cannot fully assess models' capabilities in multi-turn ESC dialogues due to none self-generated bias.
(2) Manual evaluations~\cite{liu2021towards,zheng2023building} employ human evaluators to simulate conversations between the model and users with specific distress (middle panel in Figure~\ref{F_intro}). This method requires the collection of both human-AI dialogues and manual judgments, resulting in challenges such as high cost and low efficiency.

To alleviate the above issues, we propose ESC-Eval, which replaces human labor with role-playing LLMs (right panel in Figure~\ref{F_intro}) to achieve efficient and comprehensive ESC evaluation.
We assign role-playing LLM engaging in multi-turn conversations with ESC chatbots under evaluation and collect the conversation data as the target of evaluation.
In this manner, ESC-Eval is expected to efficiently achieve performance comparable to human evaluation that involves naturalistic multi-turn dialogues data, while getting rid of reliance on ground truth and heavy labor requirements.
However, to ensure the effectiveness of our evaluation framework, two  components are important:
\romannumeral1) diverse role cards sourced from a variety of troubled individuals in real-world scenarios, which could be used to guide the LLM role-playing during evaluation and ensure the comprehensive evaluation.
\romannumeral2) A role-playing agent that closely mirrors real human behavior, enabling the acquisition of data that faithfully reflects real human interactions, thereby guaranteeing the objectivity and fairness of the evaluation results.

To accomplish these two objectives, firstly, we propose to reconstruct role cards from seven existing QA and dialogue datasets~\cite{qiu2023smile,liu2021towards,zheng2023building,sharma2020computational,lahnala2021exploring,sun2021psyqa,liu2023chatcounselor}, which are relevant to emotional companionship or psychological counseling.
However, these datasets do not contain user cards, thus, we use GPT-4 to extract and summarize the key information of users followed by a two-stage filtering process involving GPT-4 and human judgment. In this manner, we obtain 2,801 qualified role cards.
Secondly, to construct reliable role-playing agents, 
we propose to develop a role-playing agent for ESC-Eval.
In detail, we construct a dataset consisting of 3.5K ESC role-playing data from ESConv~\cite{liu2021towards}, ExTES~\cite{zheng2023building} and Smile~\cite{qiu2023smile}, each data appeared in the format of a role card and multi-turn dialogue.
We also enrich the data up to 14K by incorporating five existing role-playing instruction datasets.
Through fine-tuning Qwen1.5~\cite{bai2023qwen}, we develop a role-playing model called ESC-Role. Compared with existing state-of-the-art role-playing models, like GPT-4 and BaichuanNPC~\cite{yang2023baichuan}, the ESC-Role behaves more like a person encountering real-life troubles.

With the completion of ESC-Eval, considering huge amounts of human annotations, we select 655 high-quality role cards and comprehensively evaluate 14 LLMs with ESC-Role, including general AI-assistant LLMs (\emph{e.g.}, ChatGPT and Llama3~\cite{touvron2023llama}), and ESC-oriented LLMs (\emph{e.g.}, ExTes-Llama~\cite{zheng2023building}).
\begin{figure*}
\centering
    \includegraphics[width=\textwidth]{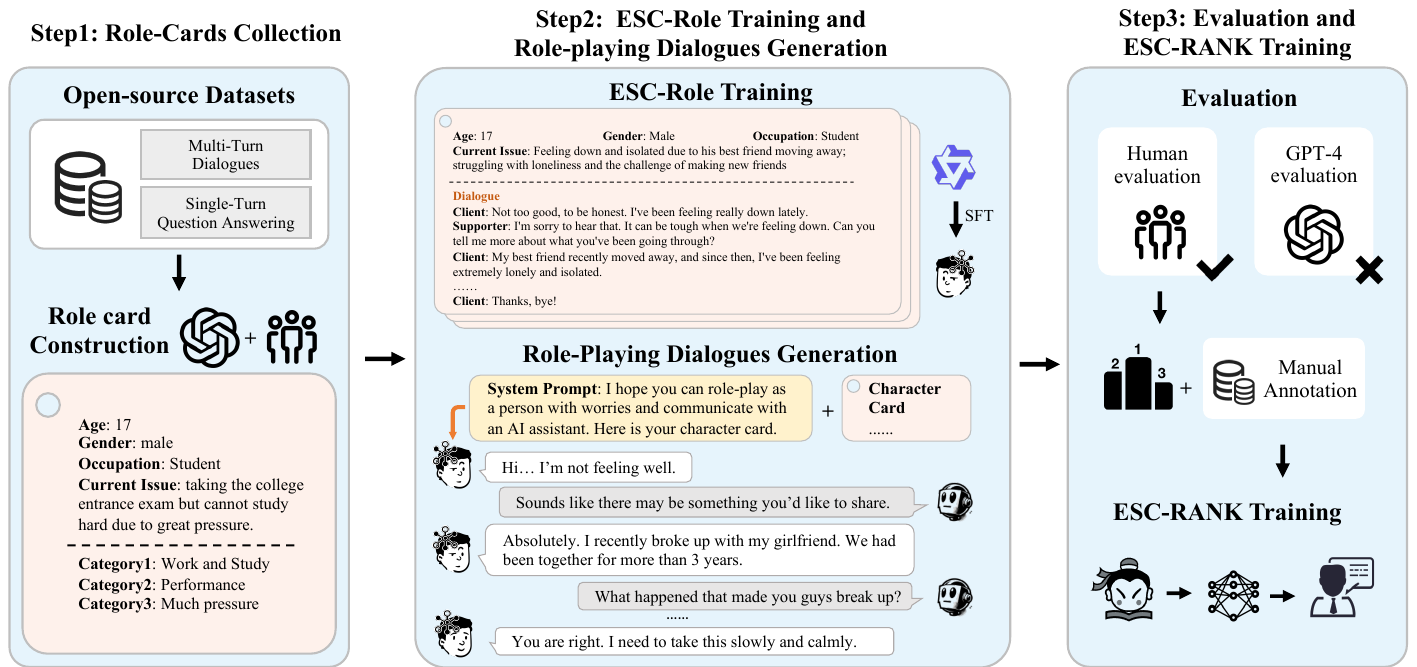}
    \caption{Overview of ESC-Eval, which used role-playing to evaluate the capability of ESC models. }
    \label{Fframework}
\end{figure*}
After obtaining 8.5K interactive dialogues based on the 14 LLMs, we conduct comprehensive human evaluations and collect 59,654 manual evaluation results in terms of 7 dimensions (\emph{i.e.}, fluency, diversity, empathy, information, humanoid, skillful, and overall). The evaluation results show that the ESC-oriented LLMs outperform most general AI-assistant LLMs, but get poor performance on emotion support knowledge and human preferences. Finally, to automate the scoring process for future ESC models, we train ESC-RANK using the 59,654 manual evaluation results, achieving a scoring performance that surpasses GPT-4 by 35 points in terms of accuracy.

Our main contributions are concluded as follows:
\begin{itemize}[leftmargin=*,topsep=0pt]
\setlength{\itemsep}{0pt}
\setlength{\parsep}{0pt}
\setlength{\parskip}{0pt}
\item We propose ESC-Eval, the first framework for evaluating LLM-based ESC models via role-playing. It features 2801 diverse user cards with fine-grained information, a dedicated role-playing model closely resembling individuals experiencing distress, and 7 meticulously designed dimensions for rigorous evaluation.  
\item Through ESC-Eval, we test 14 LLMs and manually annotate the results according to our meticulously designed dimensions. Our findings underscore an immediate demand for an ESC model exhibiting superior human preference and robust knowledge 
of emotional support.
\item For automatic evaluation of future ESC models, we developed ESC-RANK, a scoring model that outperforms GPT-4 by 35 points.
\end{itemize}

\section{ESC-Eval}
\subsection{Framework Overview}
Figure~\ref{Fframework} illustrates the workflow of ESC-Eval. ESC-Eval utilizes a role-playing model and a set of role cards to interact with ESC models under evaluation, followed by manual annotation of the obtained dialogue data. In ESC-Eval, the availability of a substantial number of diverse role cards and a more reliable role-playing agent holds paramount significance for ESC-Eval. Subsequently, the following section will outline the measures taken to ensure the reliability of these two crucial foundational components.

\subsection{Role Card Acquisition}
\label{Stwo}
To ensure the diversity of character cards, drawing inspiration from ESConv\cite{liu2021towards}, ExTES\cite{zheng2023building}, and the Life Events Scale\cite{wethington2016life}, we first construct a classification system consisting of three hierarchical layers and encompassing 37 categories. Then we propose reconstructing role cards in open-resources data and identifying each role card within each category. The construction of this procedure involves three primary steps. First, we collect 7 open-source datasets that cover a wide range of potential user roles. Then we utilize GPT-4 to extract roles from these datasets and filter out low-quality role cards, followed by human filtering. Finally, we employ a manual annotation process to ensure the quality of the role cards and classify them into their respective tertiary categories. We will introduce each step in the following, and more details can be found in Appendix~\ref{Abenchmark}.
\subsubsection{Dataset collection}
To obtain a diverse set of character cards, we conduct a comprehensive investigation into existing datasets in the field of emotion support and mental health datasets. Subsequently, we select seven datasets as the source datasets for this study. The open-source datasets utilized are listed in Appendix~\ref{Abenchmark}.
\subsubsection{User cards extraction and filtering}
 After obtaining these datasets, we encounter both Multi-turn Dialogue (MD) datasets and single-turn Question-and-Answer (QA) datasets. To extract user profiles from these diverse datasets, we employed different prompts for QA and MD datasets using GPT-4 for the initial extraction. The utilization of GPT-4 in this process incurred approximately a cost of \$120. 
 After acquiring the initial character cards, we employ GPT-4 to conduct an initial filtration process on role cards, eliminating those that solely consist of emotions without any associated events. The utilization of GPT-4 in this process incurred an approximate cost of \$70. The prompt used in this section can be found in Appendix~\ref{Abenchmark}. After the filtering process of GPT-4, we apply a human filter to ensure the quality of these cards.
\subsubsection{Manual annotation and correction}
After obtaining role cards that had undergone preliminary screening by GPT-4 and human filter, we employ a two-stage approach involving crowdsourced annotation followed by manual correction to ensure the quality of the role cards. \newline
\textbf{Crowd workers annotation}
In the above section, we develop a comprehensive three-tier classification system comprising a total of 37 categories of real-life questions, which are listed in Table~\ref{Tthreedis}. Based on this classification, the crowd workers are instructed to annotate the filtered character cards with their corresponding tertiary classifications. This requirement is motivated by the need for convenient evaluation and the quality of role-playing. Additionally, we request the crowd workers to label the high-quality and medium-quality character cards within the dataset. The annotation rules and classification system for annotation can be found in Appendix~\ref{Abenchmark}. The estimated duration for the annotation phase required is approximately 10 days with 10 crowdsourcing workers. 
\newline
\textbf{Human correction}
Upon completion of the first-stage crowdsourced annotation, we proceed with a second-stage manual correction. We request authors of this paper who are more familiar with this task to conduct an examination of the annotations for each role card, rectifying any incorrect categorizations and addressing issues pertaining to the quality of the role cards

Following the two-stage process of crowdsourced annotation and manual correction, the role cards representing various real-world individuals with different problems are successfully reconstructed. The data analysis of these role cards is listed in Appendix~\ref{Abenchmark}.
\subsection{ESC-Role}
To construct a more robust role-playing model, we train a specific role-playing agent called ESC-Role using both general data and data specific to ESC scenarios for ESC-Eval. The following sections outline the steps involved in training and evaluating this model.
\subsubsection{Data Collection}
Using the same procedure as in Section~\ref{Stwo}, we select Smile, ESConv, and ExTES datasets mentioned previously to collect ESC scenarios data. We employ methods including extraction through GPT-4, filtering with GPT-4, and manual filtering to extract role cards from multi-turn dialogues, Resulting in a total of 3,390 role-playing data which consist of a role card and a corresponding dialogue. The role cards are used as system prompts for model training. To further enhance the model's robust role-playing ability, we filter five role-playing datasets consisting of multi-turn dialogues from Huggingface\footnote{https://huggingface.co/}. After processing, we acquire 14K role-playing data instances, consisting of both general role-playing instruction data and ESC role-playing data.
\subsubsection{Implementation and Evaluation Metric}
Due to the inclusion of both English and Chinese in the character cards, we select Qwen1.5-14B-Chat as our base model. We adopt LoRA\cite{hu2021lora} parameter-efficient Finetuning on the dataset collected above.
We compare ESC-Role with some state-of-the-art role-playing agents like GPT-4 and BaichuanNPC, these agents are API-based LLMs, and we conduct all kinds of prompts like Chain-of-Thought (CoT)\cite{wei2022chain} and In-Context-Learning (ICL)\cite{min2022rethinking}, more details can refer to Appendix~\ref{Amodel}.
\begin{table*}[h]
    \centering
    \scalebox{0.85}{
        \begin{tabular}{c|p{1.25cm}p{1.25cm}p{1.25cm}p{1.25cm}p{1.25cm}p{1.25cm}|p{1.25cm}}
        \hline
        Model & Cohe. & Flue. & Them. & Comp. & Emot. & Huma. & Aver. \\
        \hline
        GPT-4$_{zero\_shot}$ & 9.9/\textbf{9.8} & 7.3/7.6 & \textbf{10}/\textbf{10} & \textbf{10}/\textbf{10} & 3.2/6.2 & 2.2/6.9 & 7.1/8.4\\
        GPT-4$_{ICL}$ & 9.9/\textbf{9.8} & 7.9/7.9 & \textbf{10}/\textbf{10} & \textbf{10}/\textbf{10} & 5.5/\underline{8.0} & 4.7/8.0 & 8.0/\underline{9.0}\\
        GPT-4$_{CoT}$  &  \textbf{10}/9.1 & 8.3/7.2 & \textbf{10}/9.2 & \textbf{10}/9.2 & 4.9/7.8 & 5.3/\underline{8.5} & 8.1/8.5 \\
        GPT-4$_{ICL+CoT}$  & \textbf{10}/\textbf{9.8} & \underline{8.9}/8.0 & \textbf{10}/\textbf{10} & \textbf{10}/\textbf{10} & 4.7/7.9 & 4.9/7.9 & 8.1/8.9 \\
        \hline
        Baichuan-NPC$_{zero\_shot}$ & 9.7/9.5 & 8.7/8.0 & 9.7/9.4 & 9.6/8.0 & \underline{6.3}/6.1 & 5.3/5.5 & 8.2/8.0\\
        Baichuan-NPC$_{ICL}$  & 9.7/9.6 & 8.5/\underline{9.1} & 9.6/9.3 & 9.3/8.3 & 5.3/5.3 & 4.7/4.5 & 7.8/7.7 \\
        Baichuan-NPC$_{CoT}$  & 9.8/9.1 & \underline{8.9}/5.9 & \textbf{10}/8.9 & 9.9/8.5 & 5.9/6.1 & \underline{6.5}/8.1 & \underline{8.5}/8.1 \\
        Baichuan-NPC$_{ICL+CoT}$  & 9.6/9.2 & 8.4/8.0 & 9.4/8.3 & 9.4/8.1 & 5.3/5.9 & 4.6/5.1 & 7.8/7.4\\
        \hline
        ESC-Role & \textbf{10}/\textbf{9.8} & \textbf{9.8}/\textbf{9.7} & \textbf{10}/\textbf{10} & \textbf{10}/9.5 & \textbf{7.5}/\textbf{9.3} & \textbf{6.6}/\textbf{9.1} & \textbf{9.0}/\textbf{9.6} \\
        \hline
        \end{tabular}
    }
    \caption{Human judgement \textbf{ZH/EN} results of different role-playing agents. }
    \label{Trole_result}
\end{table*}
To compare the effectiveness of different role-playing models, we draw upon research on role-playing and the distinctive features of the emotional support domain. We propose six categories of metrics, including general metrics (\emph{i.e.}, Coherence, Fluency) and domain-specific metrics (\emph{i.e.}, Thematic consistency, Completeness, Emotional Congruence, Humanoid, Coherence, Fluency). We use a manual evaluation method to rate each dimension on a 3-point scale.
We also conduct pairwise comparisons through manual evaluation, where human evaluators determine which dialogues resemble human-human conversations more closely. The more frequently a particular model is selected by the evaluators, the better its performance is considered to be.
\subsubsection{Evalution Results}
\begin{figure}
\centering
\includegraphics[width=0.45\textwidth]{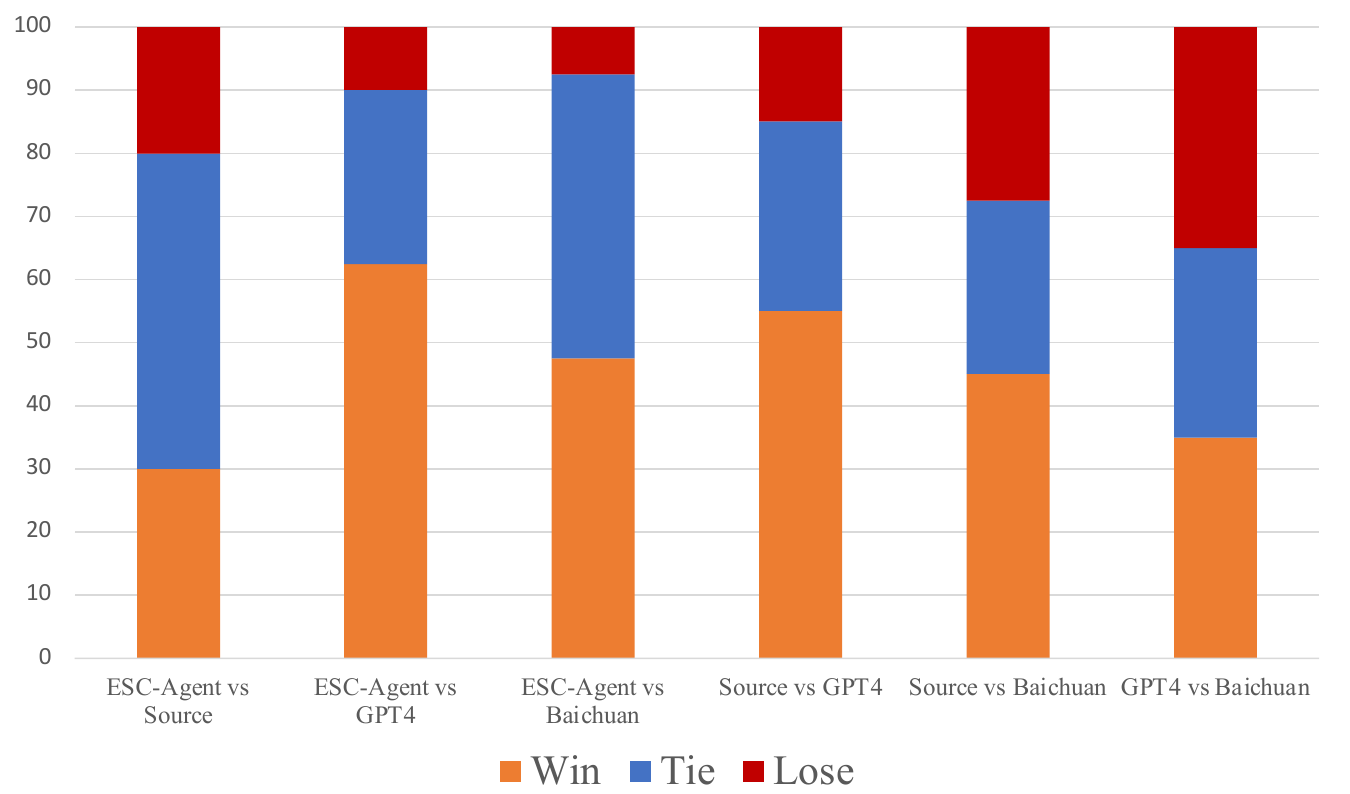}
\caption{Win rate of different role-playing agents and source data, where source denotes human dialogue. }
\label{Fwin_rate}
\end{figure}
The human judgment results of these models are presented in Table~\ref{Trole_result}. From the table, it can be observed that, in terms of the comparison of general API models, GPT-4 performs better in English, while Baichuan-NPC performs better in Chinese. The performance of GPT-4 in role-playing can be improved by optimizing various prompts, whereas Baichuan-NPC even experiences a decrease in performance with prompt optimization. Analyzing the reasons behind this, Baichuan-NPC is invoked through parameter settings\footnote{https://platform.baichuan-ai.com/docs/npc}, and it is unclear what internal strategies are employed to concatenate CoT and ICL into prompts. On the other hand, GPT-4 prompts are independently constructed by the author of this paper, which enhances its performance. Furthermore, when compared to ESC-Role, the trained ESC-Role not only demonstrates stronger human-like attributes in ESC's domain-specific metrics but also shows impressive results in genetic metrics.

In addition, we select pairs that had the best performance with different role-playing models and the source multi-turn dialogue data of role cards. We manually evaluate which dialogue more closely resembled real human conversations. The results are shown in Figure~\ref{Fwin_rate}. From the figure, it can be observed that it is difficult for humans to distinguish between the results generated by ESC-Role and the results from the original data. Both of them outperform GPT-4 and Baichuan-NPC, demonstrating the effectiveness of using ESC-Role for role-playing in ESC-Eval. 
\section{Evaluation}
In this section, we conduct evaluations on 14 general LLMs and domain-specific LLMs on ESC-Eval. We first introduce the models for evaluation. Then we display our experimental results. Finally, we display the details of our scoring model ESC-RANK.
\begin{table*}[]
    \centering
    \scalebox{0.75}{\begin{tabular}{c|cc|ccccccc|c}
    \toprule
    \multicolumn{3}{c|}{Model} & Fluency & Expression & Empathy & Information & Skillful&  Humanoid& Overall & Average \\
        \hline
        \hline
        \multirow{14}{*}{EN} & \multirow{2}{*}{Close} & GPT-4 & 74.32 & \textbf{71.68} & 71.22 & \textbf{73.72} & \textbf{74.92} & 36.40 & 44.18 & 63.78 \\
         & & ChatGPT & \underline{74.70} & \underline{71.22} & \underline{72.12} & \underline{73.19} & \textbf{74.92} & 37.08 & 45.24 & \underline{64.07} \\
        \cline{2-11}
        &\multirow{5}{*}{Open} & Vicuna-7B-1.5 & 63.37 & 67.07 & 71.00 & 71.53 & 71.68 & 41.31 & 38.67 & 60.66 \\
         & & WizardLM2-7B-Chat & 53.10 & 65.79 & 71.83 & 73.87 & 71.37 & 25.08 & 33.46 & 56.36 \\
         & & Qwen1.5-7B-Chat & 72.89 & 69.34 & 70.47 & 73.19 & 74.85 & 27.49 & 42.37 & 61.51 \\
         & & Chatglm3-6B & 74.02 & 67.82 & 70.69 & 71.37 & 74.32 & 41.84 & 42.60 & 63.24 \\
         & & Yi-6B-Chat & \textbf{75.15} & 66.99 & 69.11 & 70.39 & 71.98 & 38.82 & 43.05 & 62.21 \\
         & & LLaMa3-8B-Instruct & 63.59 & 67.37 & \textbf{72.65} & 71.90 & 74.69 & 40.55 & 41.84 & 61.80 \\
        \cline{2-11}
       & \multirow{6}{*}{Domain} & ChatCounselor & 74.54 & 66.61 & 69.03 & 64.95 & 69.03 & \underline{65.18} &\textbf{47.50} & \textbf{65.27} \\
        &  & MindChat & 74.40 & 57.85 & 67.60 & 56.80 & 61.25 & 61.71 & 39.05 & 59.81 \\
        &  & SoulChat & 25.53 & 60.20 & 66.77 & 56.27 & 60.88 & 61.25 & 36.86 & 52.54 \\
        &  &  EmoLLM & 36.56 & 68.96 & 70.85 & 71.45 & 74.47 & \textbf{65.26} & \underline{46.53} & 62.01 \\
        &  & MeChat & 52.42 & 61.10 & 66.01 & 57.63 & 61.86 & 62.01 & 39.43 & 57.21 \\
         & & ExTES-LLaMa & 74.32 & 59.97 & 69.94 & 57.02 & 62.69 & 63.52 & 41.01 & 61.21 \\
         \hline
        \hline
        \multirow{12}{*}{ZH}  & \multirow{2}{*}{Close} & GPT-4 & 71.53 & 63.97 & 64.74 & 69.14 & \underline{75.93} & 28.01 & 39.51 & 58.97 \\
         & & ChatGPT & 74.54 & 68.98 & 69.14 & \underline{70.06} & 72.38 & 32.79 & 42.75 & 61.52 \\
        \cline{2-11}
      & \multirow{5}{*}{Open} & Vicuna-7B-1.5 & 52.85 & 63.27 & 65.43 & 68.06 & 64.51 & 35.41 & 30.32 & 54.27 \\
         & & WizardLM2-7B-Chat & 54.32 & 64.04 & 66.90 & 69.75 & 65.28 & 26.08 & 30.94 & 53.90 \\
         & & Qwen1.5-7B-Chat & 74.23 & \underline{70.37} & 70.14 & 69.83 & 74.07 & 28.16 & 41.90 & 61.24 \\
         & & Chatglm3-6B & 73.53 & 67.82 & 66.74 & 68.83 & 69.44 & 27.01 & 39.35 & 58.96 \\
        &  & Yi-6B-Chat & 74.00 & 67.59 & 65.59 & 68.13 & 70.52 & 29.01 & 38.97 & 59.12 \\
        \cline{2-11}
       & \multirow{5}{*}{Domain} & ChatCounselor & 71.91 & 66.05 & 68.83 & 67.13 & 69.37 & 63.35 & 46.45 & 64.72 \\
         & & MindChat & 75.39 & 64.12 & 69.37 & 66.44 & 68.90 & 67.13 & 47.53 & 65.55 \\
        &  & SoulChat & \underline{76.16} & 65.28 & \underline{71.30} & 67.28 & 70.37 & \textbf{69.06} & \underline{48.53} & \underline{66.85}\\
         & &  EmoLLM & \textbf{78.09} & \textbf{71.45} & \textbf{74.77} & \textbf{73.15} & \textbf{78.63} & \underline{68.67} & \textbf{57.10} & \textbf{71.69} \\
         & & MeChat & 74.85 & 63.04 & 68.67 & 64.27 & 67.75 & 66.59 & 45.45 & 64.37 \\
    \toprule
    \end{tabular}}
    \caption{Human evaluation results of different models.}
    \label{Thuman_en}
\end{table*} 
\subsection{Evaluating models}
We select 14 models for evaluation, including closed-source, open-source, and domain-specific models, which are as follows: 
\begin{enumerate}
\item \textbf{Closed-source: }GPT-4~\cite{achiam2023gpt}; ChatGPT.
\item \textbf{Open-source: }Vicuna~\cite{zheng2023judging}; llama3~\cite{touvron2023llama}; WizardLM\cite{xu2023wizardlm}; Qwen1.5~\cite{bai2023qwen}; Chatglm3~\cite{zeng2022glm}; Yi~\cite{ai2024yi}.
\item \textbf{Domain-specific: }ExTES-llama~\cite{zheng2023building}; ChatCounselor~\cite{liu2023chatcounselor}; MindChat~\cite{MindChat}; SoulChat~\cite{chen2023soulchat}; EmoLLM~\cite{EmoLLM}; MeChat~\cite{qiu2023smile}.
\end{enumerate}
To facilitate a more accurate comparison of the capabilities of various models, we choose models of similar magnitudes, such as the 6B/7B/8B model parameter sizes for comparison. 

\subsection{Evaluation Results} 

Based on pre-defined dimensions, we conduct a comprehensive manual assessment, and the results are presented in Table~\ref{Thuman_en}. Both in English and Chinese ESC conditions, domain-specific LLMs (ChatCounselor and EmoLLM), respectively achieved the best results.
From Table~\ref{Thuman_en} above, in the comparison between general models and domain-specific models, the general models perform better in terms of fluency, expression diversity, and emotional comfort skills. This can be attributed to their highly structured output, such as phrases like ``I understand you very well, it is very normal to feel \ldots, here are some possible suggestions: '' The general models generate a large amount of text, scoring high in terms of advice effectiveness and expression diversity. Besides, due to Larger scale parameters, the API-based models exhibit greater knowledge of emotional comfort, with GPT-4 and ChatGPT demonstrating the highest proficiency. However, these models perform poorly in terms of human-like and human-centric responses, as users in this context expect replies that are more humanized and possess greater human-like qualities. In the comparison of domain-specific models, MindChat, SoulChat, and EmoLLM, which are not fine-tuned in English, showed inferior fluency. On the other hand, ExTES-llama and ChatCounselor perform well. ExTES is fine-tuned with data generated by ChatGPT, while ChatCounselor is fine-tuned using real psychological counseling data, exhibiting superior performance.
From Table~\ref{Thuman_en} bottom, the general models perform well in terms of expressing diversity and providing effective suggestions. Trained on diverse and abundant data, EmoLLM exhibits excellent performance across multiple dimensions among the various domain-specific models. Other domain-specific models, due to their remarkable human-like qualities and human convenience, surpass the general models. However, there is still room for improvement in terms of emotional support knowledge, and significant potential exists for enhancing human convenience. It is worth noting that MindChat, trained on bilingual data, not only demonstrates strong Chinese language proficiency but also exhibits commendable English language capabilities.

\subsection{Correlation Analysis}
\begin{table*}[]
    \centering
    \scalebox{0.85}{\begin{tabular}{c|cccccccccccc}
    \toprule
    \multirow{2}{*}{\textbf{Metrics}} & \multicolumn{2}{c}{Fluency} &\multicolumn{2}{c}{Suggestion} & \multicolumn{2}{c}{Skillful}  & \multicolumn{2}{c}{Empathy} & \multicolumn{2}{c}{Overall} & \multicolumn{2}{c}{Average} \\
    & Spear. & Pear. & Spear. & Pear.  & Spear. & Pear. & Spear. & Pear. & Spear. & Pear.  & Spear. & Pear. \\
        \hline
         Bleu-1 & \textbf{40.60} & 40.63 & -67.20  & -65.68   & -51.68 & -51.00  & -28.32 & -27.53  & -55.92 & -52.95 & -60.98 & -56.36  \\
         Bleu-2 & 18.16 & 12.05  & -18.82  & -15.81   & -2.97 & -0.29  & -29.34 & -22.91  & -21.66 & -19.97  & -18.25 & -17.95  \\
         Bleu-4 & -0.04 & -2.56  & -5.40  & -3.33  & \underline{27.54} & \underline{22.97} &  -0.90 & -14.38 & \underline{13.50} & \underline{2.99}  & 10.78 & 2.80  \\
         Distinct-1 & 37.92 & \textbf{43.84}  & -79.61  & -81.95   & -59.52 & -62.17  & -36.18 & -32.60  & -62.47 & -65.36  & -70.02 & -68.11  \\
         Distinct-2 & \underline{38.63} & \textbf{43.84}  & -81.51  & -80.79   & -61.07 & -61.45  & -37.09 & -36.21  & -65.46 & -65.32 & -72.53 & -69.67  \\
         Rouge-L & 38.25 & 36.77  & -56.98  & -58.27  & -36.03 & -37.31  & -19.23 & -23.05 &  -42.80 & -45.31  & -45.22 & -45.59  \\
        Meteor & 8.01 & 14.94  & \underline{20.09}  & \underline{12.76}  & 1.23 & -0.34 &  \textbf{14.31} & \textbf{10.11}  & 6.77 & 0.97 & \underline{17.30} & \underline{13.73}  \\
        \hline
        ESC-Eval & -1.61 & -0.69 &  \textbf{36.26}  & \textbf{33.36}  & \textbf{39.02} & \textbf{38.70} & \underline{9.17} & \underline{6.02} & \textbf{45.01} & \textbf{44.58}  & \textbf{46.31} & \textbf{46.05} \\
    \toprule
    \end{tabular}}
    \caption{Sample-level Spearman correlation (Spear.) correlation and Pearson (Pear.) correlation of different metrics.}
    \label{Tcorrlation}
\end{table*}
To validate the effectiveness of ESC-Eval, we randomly select 20 instances from the ESConv dataset. We choose three categories from the target model and included five different models for 
correlation analysis. These models are subjected to interactions with human evaluators who model seekers looking for help. And they are asked to provide ratings upon completion of the interactions, according to human evaluation methods in other papers. The human-rated scores are considered as the optimal evaluation method, and we conduct a correlation analysis between various automatic evaluation methods and the ESC-Eval method. The results are presented in Table~\ref{Tcorrlation}.
\begin{table}[]
    \centering
    \scalebox{0.87}{\begin{tabular}{cc|cc}
    \toprule
    Dim. & Model &  ACC & ACC$_{soft}$ \\
        \hline
        \multirow{3}{*}{Flu.} & InternLM2 & 31.84/17.15 & 93.07/72.82 \\
        & GPT-4 & 35.82/25.89 & 95.51/89.21 \\
        & ESC-RANK & 88.45/81.66 & 99.87/99.24 \\
        \hline
        \multirow{3}{*}{Exp.} & InternLM2 & 27.09/26.21 & 54.94/56.09 \\
        & GPT-4 & 60.59/66.02 & 96.53/99.57 \\
        & ESC-RANK & 65.72/68.39 & 99.49/99.67 \\
        \hline
        \multirow{3}{*}{Emp.} & InternLM2 & 19.38/14.56 & 80.74/84.90 \\
        & GPT-4 & 41.46/48.11 & 88.58/94.28 \\
        & ESC-RANK & 69.70/77.02 & 99.10/98.71 \\
        \hline
        \multirow{3}{*}{Inf.} & InternLM2 & 35.94/32.58 & 83.83/88.03 \\
        & GPT-4 & 56.35/68.28 & 94.22/98.27 \\
        & ESC-RANK & 75.10/77.02 & 98.97/99.46 \\
        \hline
        \multirow{3}{*}{Ski.} & InternLM2 & 32.34/27.5 & 84.85/91.15  \\
        & GPT-4 & 27.98/38.83 & 82.03/91.80 \\
        & ESC-RANK & 79.72/68.61 & 96.79/99.57 \\
        \hline
        \multirow{3}{*}{Hum.} & InternLM2 & 22.85/25.89 & 52.25/66.77 \\
        & GPT-4 & 1.02/3.02 & 32.48/35.06 \\
        & ESC-RANK & 57.51/70.77  & 98.84/98.17 \\
        \hline
        \multirow{3}{*}{Ove.} & InternLM2 & 8.04/6.04 & 48.27/46.28 \\
        & GPT-4 & 1.80/1.73 & 15.15/17.04 \\
        & ESC-RANK & 57.89/55.45  & 99.49/99.35 \\
        \hline
        \multirow{3}{*}{Avg.} & InternLM2 & 25.50/21.42 & \underline{79.59/76.56} \\
        & GPT-4 & \underline{32.15/35.98} & 72.07/75.03 \\
        & ESC-RANK & \textbf{70.53/71.27} & \textbf{98.93/99.17} \\
        \hline
    \toprule
    \end{tabular}}
    \caption{Scoring performance comparation, while ACC denotes accuracy, ACC$_{soft}$ denotes one point deviation. }
    \label{Tesc_rank}
\end{table}
From Table~\ref{Tcorrlation}, it can be observed that ESC-Eval exhibits the best correlation with the evaluation metrics, except for the Fluency and Empathy indicators. In terms of Fluency, automated metrics outperform ESC-Eval, we analyze that during manual annotation, human annotators may exhibit some bias towards the fluency of segmented statements generated by a general model which significantly deviates from the ESConv dataset. It is observed that human annotators tend to prefer naturally expressed content, leading to relatively lower manual ratings for the fluency outputs of general models. At the same time, the content generated by the general model is quite different from that of ESconv, and the automation metric is also very low. As a result, there is a strong correlation between automated evaluation metrics and humans. However, in ESC-Eval all models perform well on fluency due to the capability of LLMs, leading to low correlation. A similar phenomenon is observed for the Empathy indicator, where although there is some correlation, it is due to the alignment process that most LLMs undergo, which enables them to display decent comforting abilities and analytical skills. In terms of the overall average metric, ESC-Eval demonstrates the most significant correlation compared to the automated metrics, further emphasizing the effectiveness of ESC-Eval. More correlation experimental results are in Appendix~\ref{Aevaluation}.
\subsection{ESC-RANK}
To facilitate subsequent research, based on InternLM2-7B-Chat\cite{cai2024internlm2} and using the manually annotated data in this article, we train ESC-RANK. ESC-RANK can score the results of multiple rounds of dialogues of different models to our well-designed dimension. 

We randomly divide the annotated data into a training set, validation set, and test set according to 7:1:2. Compared with the base model and GPT-4, the results are shown in Table~\ref{Tesc_rank}.

From Table~\ref{Tesc_rank}, it can be observed that ESC-RANK demonstrates the best scoring capability, surpassing GPT-4 by 35 points in terms of accuracy. As human scoring may not always have a clear-cut boundary, a tolerance of one-point error is allowed in scoring which denotes the result of ACC$_{soft}$. When considering ACC$_{soft}$, ESC-RANK achieves an accuracy rate of over 99\%, providing a solution for subsequent automation processes. Interestingly, GPT-4 performs poorly in the dimension of humanoid and human preference scoring. The analysis suggests that GPT-4 assigns higher scores to its own generated content or content similar to its own, which can be easily judged during human evaluation, particularly in formatted outputs such as bullet-point suggestions, where it becomes apparent that the content is machine-generated, leading to a poor score of humanoid and human preference. InternLM2 also has the same problem in human preference behavior, but it performs better in humanoid scoring, which leads to higher performance than GPT-4 in ACC$_{soft}$.

\section{Related Work}
\subsection{Emotion Support Conversation}
Traditional research~\cite{sharma2020computational,medeiros2018using,rashkin2018towards} on emotion support systems initially focus on simple single-turn emotion dialogue systems. With the emergence of the ESConv~\cite{liu2021towards} dataset, the development of ESC shifted towards more complex multi-turn dialogues. 
Researchers have proposed various optimization strategies on ESConv dataset, \citet{peng2022control} introduced an innovative hierarchical graph network, aiming to effectively utilize both the global emotion cause and the local user intention in emotional support conversations. Moving away from relying on a single strategy for response generation, \citet{tu2022misc} incorporate commonsense knowledge and a mix of response strategies into the framework of emotional support conversation and so on.
With the development of LLMs, their generative architecture has naturally made them well-suited for chatbot scenarios.Researchers~\cite{zheng2023building,qiu2023smile,liu2023chatcounselor} utilize these models by pertaining and fine-tuning through supervised learning. For instance, \citet{zheng2023building} use ChatGPT to generate data for constructing an emotion-supported dialogue system, while \citet{madani2024steering} expand the ESconv dataset to address the issue of extrapolating the length capabilities of large language models. In addition, some studies~\cite{hua2024large,zhang2024cpsycoun,chen2023llm} also use LLMs in ESC-related fields, such as psychological counseling. The purpose of our study is to provide a comprehensive and rigorous evaluation of these LLM-based ESC models.

\subsection{Role Play Agents}
Recent advancements in LLMs have significantly boosted the rise of Role-Playing Language Agents (RPLAs)~\cite{chen2024persona}. Existing researches~\cite{wang2024rolellm,tu2024charactereval,shen2024roleeval,wang2024incharacter} have proposed multiple evaluation datasets for role-playing, wherein various approaches~\cite{li2023chatharuhi,shao2023characterllm,wang2024rolellm,zhou2023characterglm} such as In-Context-Learning (ICL)~\cite{min2022rethinking}, Chain-of-Thought (CoT)~\cite{wei2022chain} and Supervised Fine-Tuning (SFT). Additionally, the industry has witnessed the emergence of numerous role-playing products, like Character AI\footnote{https://character.ai/} and Reflection AI\footnote{https://reflectionai.xyz/}, leading to a wide-ranging impact. RPLAs are capable of assuming specific roles, engaging in human-like interactions through composite character settings, role background knowledge, and speech styles, thereby exhibiting human-like attributes and playing a role in everyday conversational contexts. This paper follows the main idea of evaluating ESC models through RPLAs.

\section{Conclusion}
This paper proposes a novel approach to evaluate the effectiveness and sustainability of the Emotion Support Conversation (ESC) in Large Language Models (LLMs) by utilizing a role-playing model to acquire multi-turn dialogue data. Experimental results demonstrate the efficacy and viability of our proposed method. Our evaluation outcomes indicate that while some ESC models currently outperform general models, there is still significant room for improvement in terms of these models' knowledge capabilities and human-preference abilities. We encourage researchers to participate in ESC research and contribute to the development of more robust ESC models.

\section*{Limitations}
The crowdsourced annotators in this article are not native English speakers, but all of them are proficient English users. We employed a total of 14 annotators. Among them, one has a background in computer science, two in mental health sciences/psychology, and the remaining 11 in humanities and social sciences such as law, sociology, and history. Three are PhD candidates in China, while the others are master's students.  However, they still cannot avoid possible shortcomings in English annotation.

The role cards published in this paper were extracted using GPT4.  As much as possible this project uses a variety of filtering methods, including model filtering and manual filtering, there is still no guarantee of potential bias in the models included in the role cards. We hope future projects can improve.


\section*{Ethical Considerations}
Since this research is related to psychology, the format of the datasets used in this article has been converted, and each data instance has been manually reviewed to confirm that there are no ethical and privacy issues in each piece of data and that it complies with legal and regulatory requirements.

\section*{Acknowledgements}
This work was supported by the National Key R\&D Program of China (2022ZD0160103), Shanghai Science and Technology Innovation Action Plan (No. 22511104700) and Shanghai Artificial Intelligence Laboratory.

\bibliography{anthology,custom}

\appendix
\section{Benchmark}
\label{Abenchmark}
\begin{figure*}
\centering
    \includegraphics[width=\textwidth]{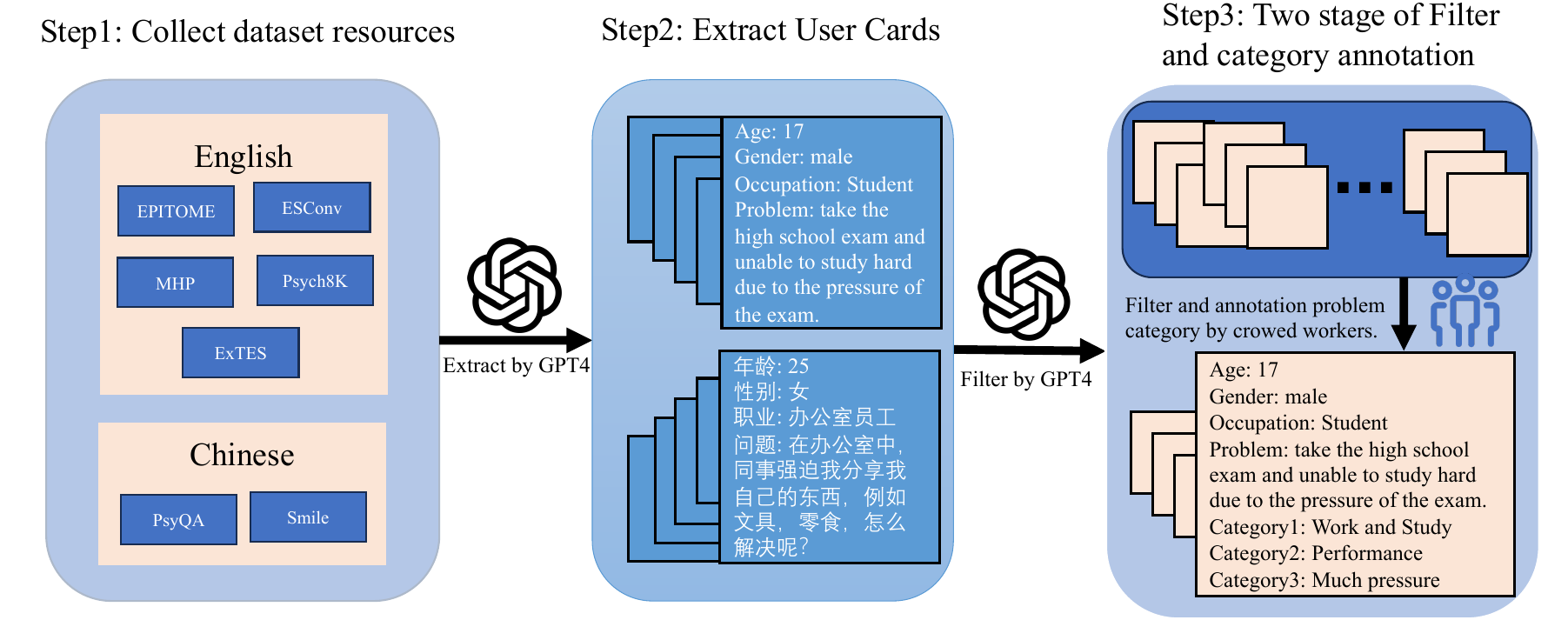}
    \caption{The framework of user-card construction. Firstly, the initial user cards are extracted from open-source datasets using GPT-4. In the second step, based on the scene classification we designed, GPT-4 is utilized to determine the category to which the character sheet data belongs, and further filtering is performed. In the third step, we employ crowdsourcing to annotate the category and subcategories of the scenes, and manually filter the user cards again.}
    \label{Fbench}
\end{figure*}

The construction of character cards, as illustrated in Figure~\ref{Fbench}, primarily consists of three steps. The first step involves collecting the raw dataset, followed by the second step of utilizing GPT-4 to extract and filter the character cards. The third step entails a two-stage manual filtering and annotation process. The following sections will provide further details on the construction procedure.

\begin{figure*}[h]
\centering
    \includegraphics[width=\textwidth]{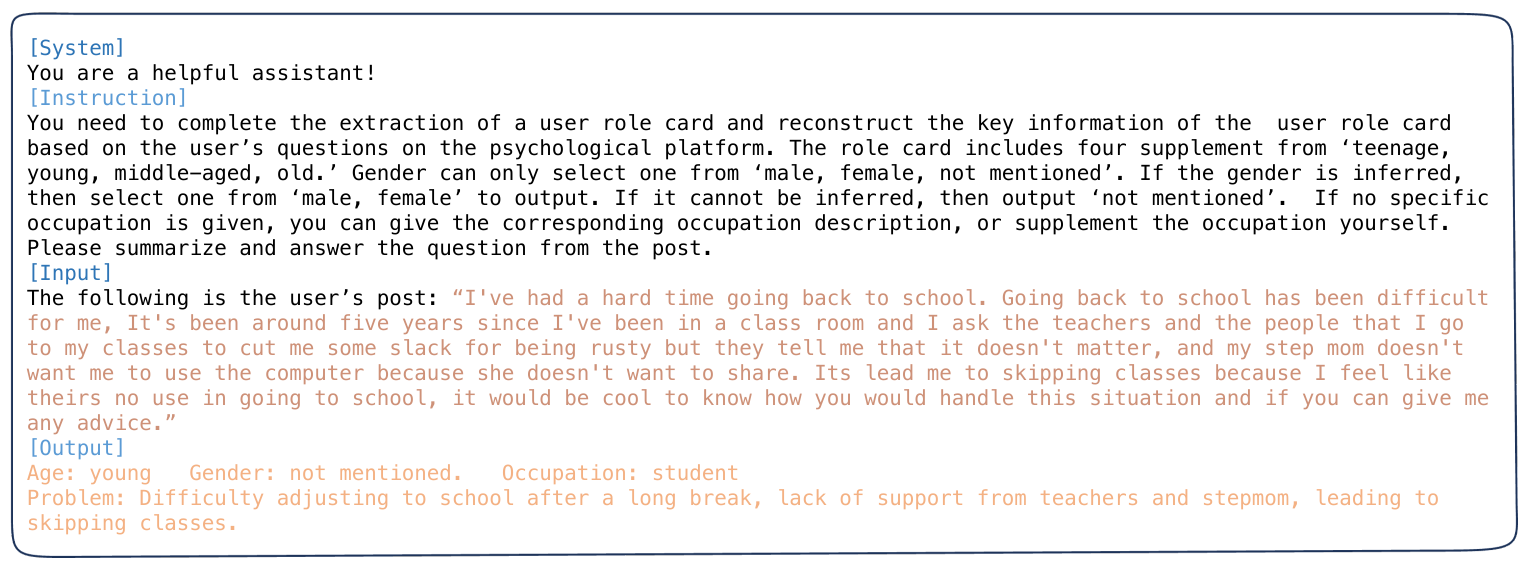}
    \caption{Prompt used for QA datasets user cards extraction. }
    \label{Fprompt_QA}
\end{figure*}

\begin{figure*}[h]
\centering
    \includegraphics[width=\textwidth]{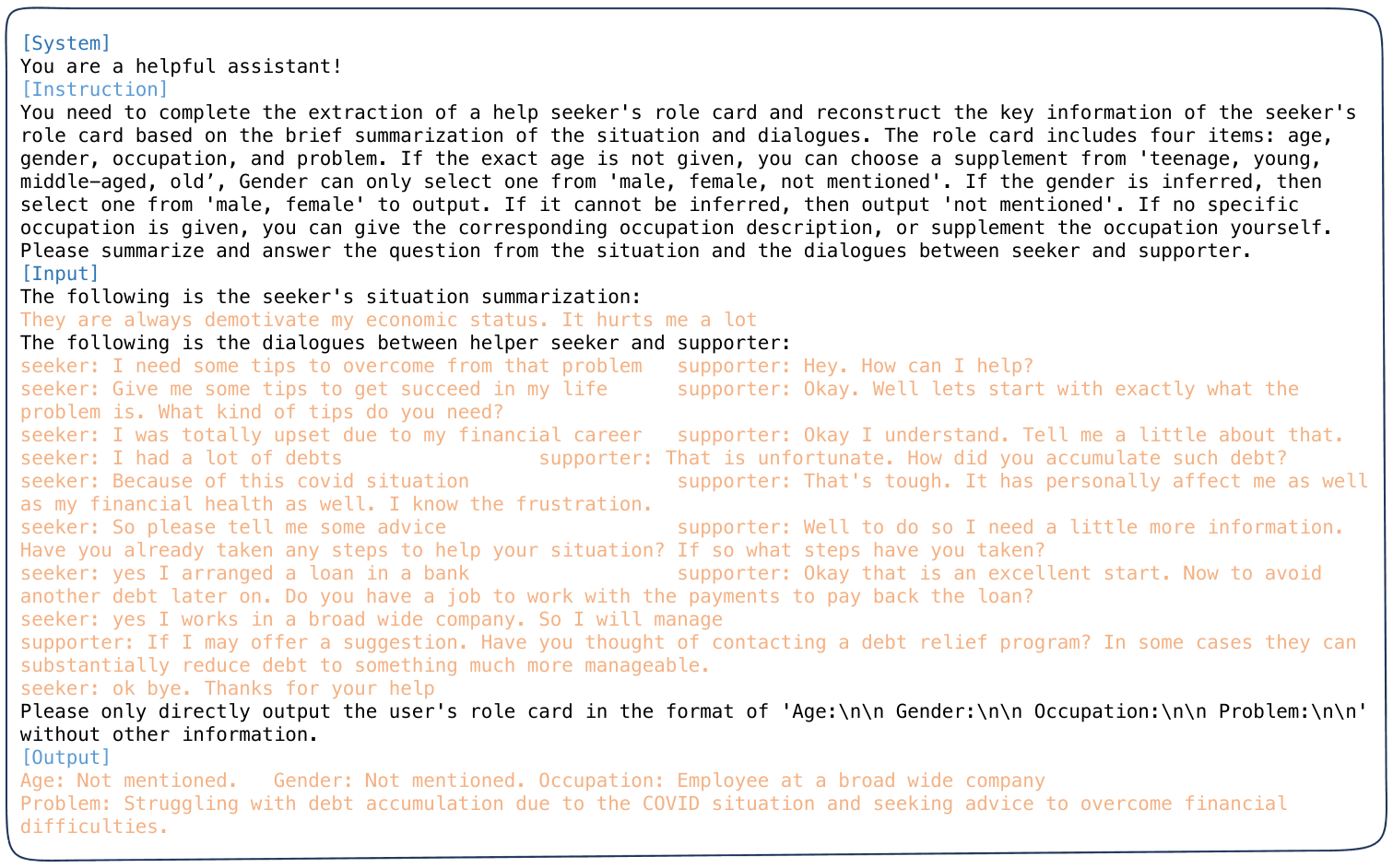}
    \caption{Prompt used for MD datasets user cards extraction. }
    \label{Fprompt_MD}
\end{figure*}

\begin{figure*}[h]
\centering
    \includegraphics[width=\textwidth]{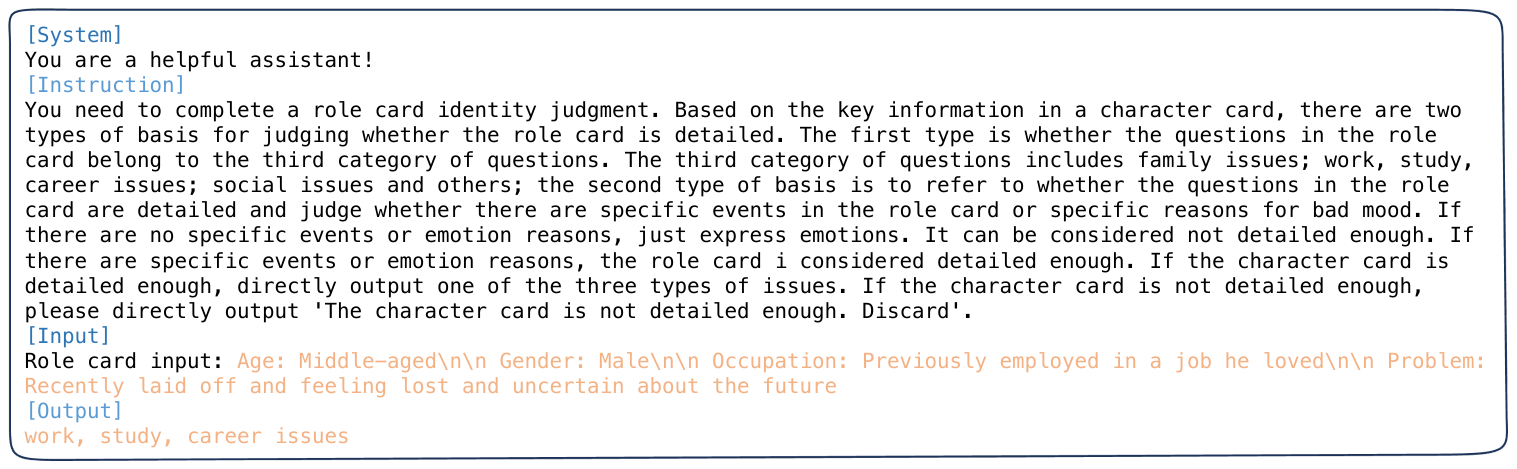}
    \caption{Prompt used for GPT-4 filtering user cards. }
    \label{Fprompt_filter}
\end{figure*}

\begin{figure*}[h]
\centering
    \includegraphics[width=\textwidth]{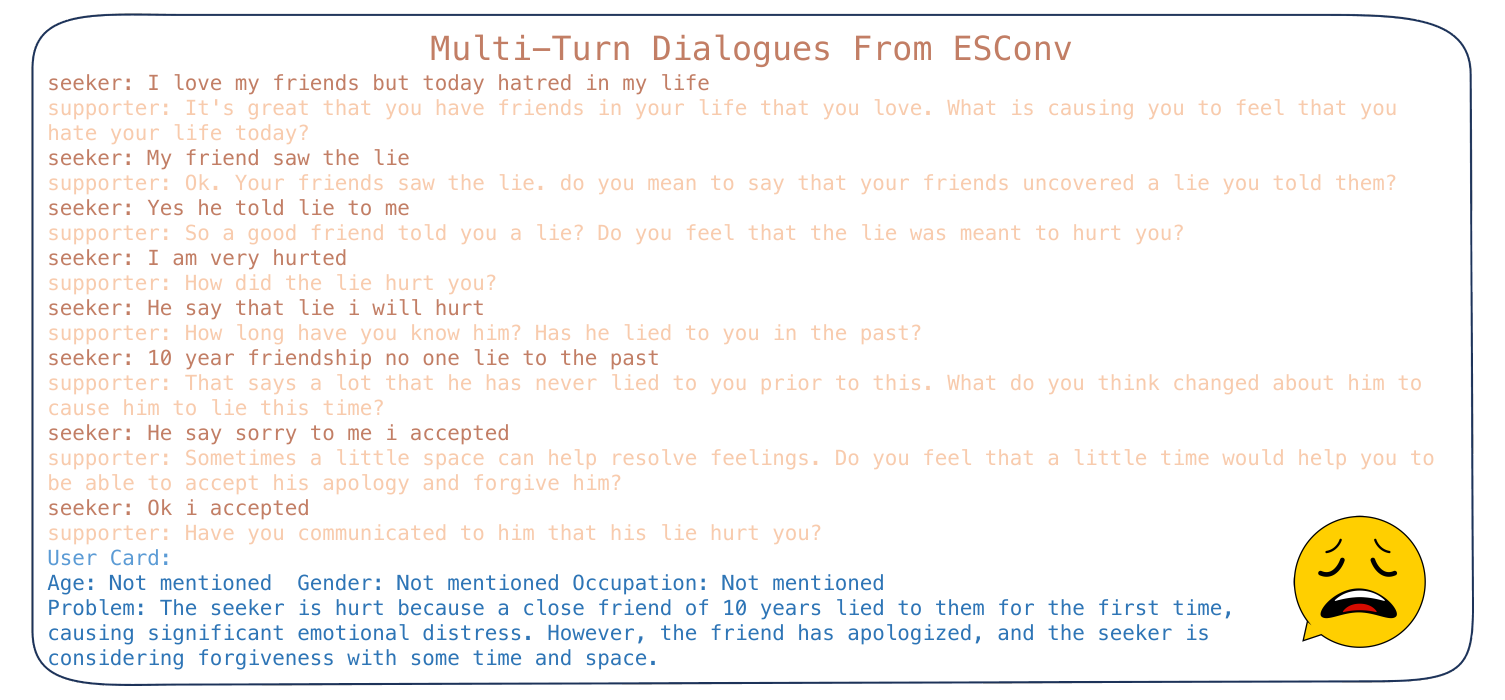}
    \caption{A case of Multi-turn dialogue which is from ESConv dataset.}
    \label{Fmd_case}
\end{figure*}

Firstly, the raw dataset used in this study can be found in Table~\ref{Tdataset}. The prompts used for extraction and filtering during the GPT-4 phase can be referenced in Figure~\ref{Fprompt_QA}, Figure~\ref{Fprompt_MD} and Figure~\ref{Fprompt_filter}. The manual annotation phase primarily relies on internal annotations within the character cards, which indicate their corresponding quality and three-tier classification. The descriptions of character cards with different quality levels are presented in Table \ref{Tthreequlity}, and the annotation guidelines for character cards are provided in Table~\ref{Trolerule}. The distribution of the completed three-tier classifications can be observed in both Table~\ref{Tthreedis} and Figure~\ref{Froledis}. Finally, we present two cases showcasing the extraction of character cards using single-turn QA and multi-turn dialogues in Figure \ref{Fqa_case} and Figure \ref{Fmd_case}.
\subsection{Resource Datasets}
\begin{table*}[h]
    \centering
    \scalebox{0.8}{
    \begin{tabular}{c|c|ccccc}
        \toprule
         \textbf{Language}& \textbf{Dataset} & \textbf{Format}  &  \textbf{Domain}  & \textbf{Resource} & \textbf{Sample\_Num} \\
        \hline
        \multirow{5}*{English} & EPITOME~\cite{sharma2020computational} & QA & Empathetic response generation  & Human & 500 \\
        ~ & MHP Reddit~\cite{lahnala2021exploring} & QA & Mental healthy counseling & Human & 1000 \\
       ~ & Psych8K~\cite{liu2023chatcounselor} & QA & Mental health support & Human &  1000 \\
       ~ & ESConv~\cite{liu2021towards} & MD & Emotional support conversation & Human & 1000\\
       ~ & ExTes~\cite{zheng2023building} & MD & Emotional support conversation & ChatGPT &500 \\
        \hline
        \multirow{2}*{Chinese} &PsyQA~\cite{sun2021psyqa} & QA & Mental healthy counseling & Human&1000 \\
        ~ & Smile~\cite{qiu2023smile} & MD & Mental healthy counseling & ChatGPT &1000 \\
        \bottomrule
    \end{tabular}}
    \caption{Open-source datasets used in our study. QA denotes datasets consisting of question-answer pairs, while MD denotes datasets consisting of multi-turn dialogues; Resource denotes whether the datasets are collected by humans or ChatGPT; Sample\_Num denotes the numbers used for user card construction in this study. }
    \label{Tdataset}
\end{table*}
\label{Adatsets}
Here are more details about these datasets:
\begin{itemize}
\item{\textbf{EPITOME:}} Psychologically relevant user post data collected on Reddit, single-turn conversation format in English.
\item \textbf{MHP Reddit:} Psychologically relevant user post data collected on Reddit, single-turn conversation format in English.
\item \textbf{Psych8K:} Real psychological consultation voice data, converted into text and processed by ChatGPT, psychologically related English multi-turn conversation data.
\item \textbf{ESConv:} Collected by crowdsourcing workers in the field of emotional dialogue, multi-turn dialogues format in English.
\item \textbf{ExTes:} Generated by ChatGPT in the field of emotional dialogue, multi-turn dialogues format in English.
\item \textbf{PsyQA:} Collected from user posts in a psychological platform, single-turn dialogue format in Chinese.
\item \textbf{Smile:} Generated by ChatGPT in the field of mental health, multi-turn dialogues format in Chinese.
\end{itemize}
These datasets encompass both single-turn question-answering and multi-turn dialogue in the domain of emotions or mental health. Through these diverse data sets, we can collect diverse real-world people who encounter a variety of problems.
\subsection{Construction Details}
The differentiation of role cards into three quality categories is based on the following two considerations: 1) Insufficient information content in ineffective role cards, resulting in a lack of specific themes for model role-playing. 2) High-quality role cards possess richer information, enabling more specific tasks and yielding more effective evaluation results.
A tripartite role card is presented in Table~\ref{Tthreequlity}, illustrating three distinct categories.

It should be noted that there is no absolute boundary between high-quality character cards, medium-quality character cards, and invalid character cards. The only difference between them lies in the richness of character information. A higher level of character richness is believed to contribute to better model performance and is more conducive to subsequent evaluations. We can only relatively identify high-quality character cards, medium-quality character cards, and invalid character cards. The boundary between valid and invalid character cards depends on whether events occurring within the characters can be classified. The classification of events is shown in the table. The boundary between high-quality and medium-quality character cards is whether events, events causes, events results, and detailed descriptions of events can be identified based on event classification. The table can be used as a reference for annotation. In the case of invalid character cards, during the dialogue between the role-playing model and the test model, only emotions are present, which exhibits redundancy among a large number of character cards. This redundancy is not conducive to simulating individuals who encounter a variety of problems in real life. Both medium-quality character cards and high-quality character cards are effective for evaluation, but high-quality character cards are more targeted.
\newline
\textbf{Human Correction rules}
To ensure the quality of collected role cards, our human correction rules are listed below:
If a crowdsourced worker deemed a role card invalid, it was discarded. If one worker classified a role card as high-quality and the other as middle-quality, a third participant corrected its classification to either high-quality or middle-quality. The remaining role cards were considered middle-quality. This process essentially involved correcting the categorization of the role cards. For the middle-quality and high-quality role cards, if two crowdsourced workers agreed on the same category, that category was accepted. If the two workers disagreed on the three-level classification, the project participant intervened to correct it, ensuring the accurate classification of the middle-quality and high-quality role cards. 
\subsection{Data Analysis}
\begin{figure*}[h]
\centering
    \includegraphics[width=\textwidth]{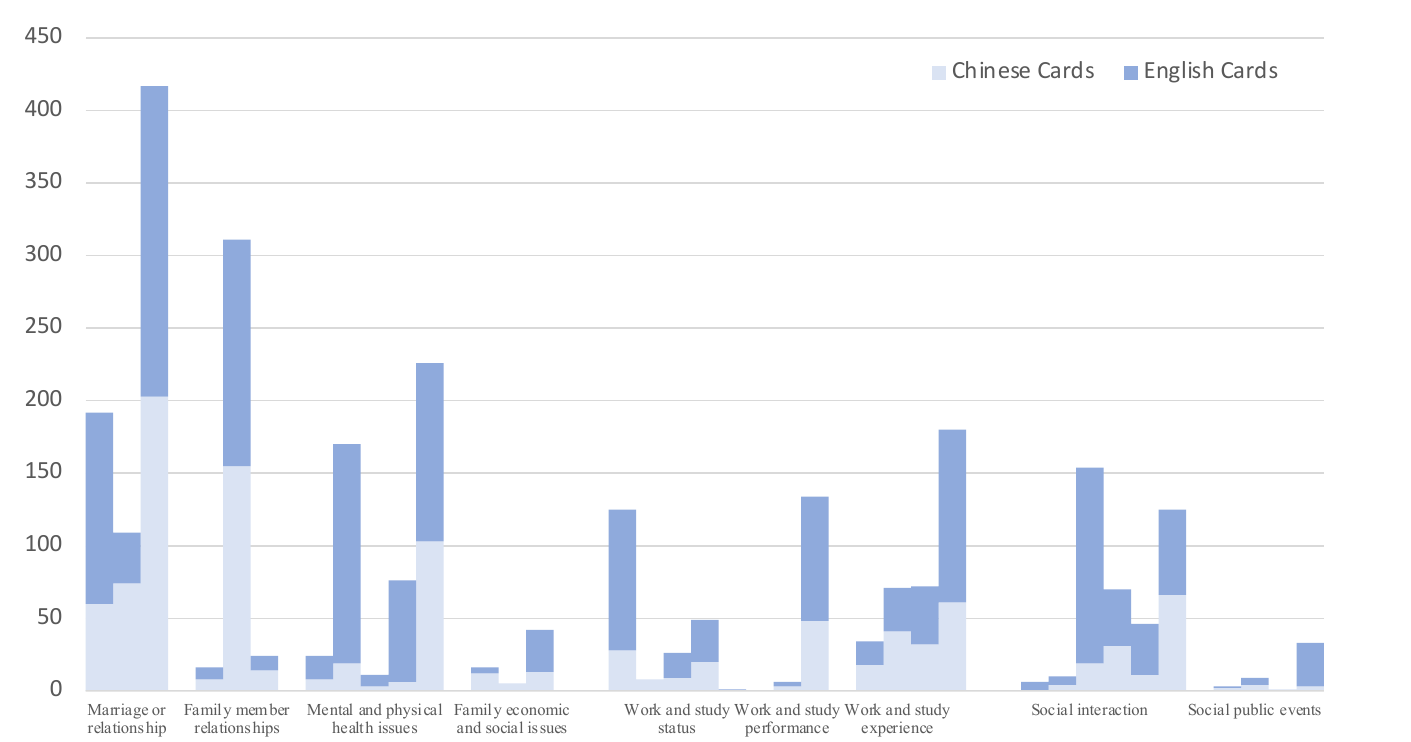}
    \caption{Role cards distribution of our constructed benchmark.}
    \label{Froledis}
\end{figure*}

\textbf{Basic analysis}
We employed a multi-step process involving both rule-based and manual methods to ensure the quality of character cards. Table~\ref{Tbasicanalysis} presents the quantities of character cards at each stage after filtration. The distribution of collected role cards is shown in Figure~\ref{Froledis}.
\begin{table}[h]
    \centering
    \scalebox{0.7}{
    \begin{tabular}{c|c|c|c|c|c}
    \toprule
        \textbf{Language} & \textbf{Extract} & \textbf{GPT-4\_F} & \textbf{Human\_F} & \textbf{High} & \textbf{Middle} \\
        \hline
        English & 3673 & 2792 & 1708 & 331 & 1455 \\
        Chinese & 2023 & 1566 & 1093 & 324 & 769 \\
    \bottomrule
    \end{tabular}
    }
    \caption{The quantities of role cards at each stage. Extract represents the initial number of role cards extracted from open resource datasets; GPT-4\_F represents the number of role cards after the filtering process using GPT-4; Human\_F represents the number of role cards after manual filtering; High represents high-quality role cards, and Middle represents medium-quality role cards.}
    \label{Tbasicanalysis}
\end{table} 
\begin{table*}[h]
    \centering
    \scalebox{0.85}{
    \begin{tabular}{|c|p{14cm}|}
    \hline
        \textbf{Type} & \textbf{Content}  \\
        \hline
        \multirow{11}*{High} &  Age: Not mentioned\qquad Gender: Not mentioned\qquad Occupation: Not mentioned
        \newline Problem: Feeling excluded and hurt by not being invited to a friend's house party, leading to feelings of loneliness and betrayal. \\
        ~ & \hrule  Age: Not mentioned\qquad Gender: Not mentioned\qquad Occupation: Resident of an apartment complex \newline Problem: The seeker reported a neighbor's dog for attacking their own dog, leading to the neighbor being evicted. Now, the seeker is facing social ostracization and blame from other neighbors, feeling like an outcast for taking action to protect their pet.\\ 
        ~ & \hrule \begin{CJK}{UTF8}{gkai}年龄：青年\qquad 性别：未提及\qquad职业：大学生
        \newline 问题：内向且怕生的大学生面临公众演讲的恐惧，担心自己在台上失声或晕倒，受到小时候校园暴力的影响，害怕人多的地方和被人注视，对自己的长相感到自卑，担心在公众场合出丑，对自己的能力和未来感到不自信。\end{CJK} \\
        \hline
        \multirow{9}*{Medium} & Age: young\qquad Gender: female\qquad Occupation: not mentioned
        \newline Problem: Confused about a male friend's feelings towards her and unsure how to proceed. \\
        ~ & \hrule Age: young\qquad Gender: female\qquad Occupation: not mentioned
        \newline Problem: The user's husband, influenced by his father, is acting differently and planning to move out of California against her wishes, and expects her to contribute all her income and time to a joint family account controlled by the men in his family. \\
        ~ & \hrule \begin{CJK}{UTF8}{gkai} 年龄：中年\qquad  性别：未提及\qquad  职业：当前职业未明确，但表明想转行至心理咨询相关领域
        \newline 问题：工作进入瓶颈期和倦怠期，面对转行至心理学领域感到焦虑和恐慌，寻求建议。 \end{CJK} \\
        \hline 
        \multirow{6}*{Invalid} & Age: not mentioned\qquad Gender: female\qquad Occupation: not mentioned \newline Problem: GI issues from metformin, considering switching to XR. \\
        ~ & \hrule Age: young\qquad Gender: not mentioned\qquad Occupation: not mentioned
        \newline Problem: anxiety and paranoia affecting relationships \\
        ~ & \hrule \begin{CJK}{UTF8}{gkai}
            年龄：中年\qquad 性别：男 \qquad 职业：未提及
            \newline 问题：最近一个月内经历失眠、焦虑、烦躁和身体不适，面临家庭压力和个人情感决策困难，导致对未来感到迷茫，有时产生极端消极想法。
        \end{CJK}\\
    \hline
    \end{tabular}
    }
    \caption{Some cases of different quality role cards.}
    \label{Tthreequlity}
\end{table*} 

\begin{table*}[h]
    \centering
    \scalebox{0.85}{
    \begin{tabular}{|c|p{14cm}|}
    \hline
        \textbf{Type} & \textbf{Rules}  \\
        \hline
        \multirow{3}*{Invalid} &  1.The character card only includes subjective emotions and thoughts, without events that elicit emotions. \newline 2. There are events present, but suitable event categorizations cannot be found, rendering the events unable to reach a granular level of classification. \\
        \hline
        \multirow{3}*{Middle} & 1. Events occur and can be classified into fine-grained categories.
 \newline 2. The causes of the events and the resulting consequences are not presented.
\newline 3. In the context of interpersonal relationships, the portrayal of the other person's image is absent.\\
        \hline
        \multirow{3}*{High} & 1. Events occur and can be classified into fine-grained categories.
 \newline 2. The causes of the events and the resulting consequences are presented.
\newline 3. In the context of interpersonal relationships, the portrayal, introduction, and description of the other person's image within the relationship are included.\\
    \hline
    \end{tabular}
    }
    \caption{The rules of three types of role cards annotations.}
    \label{Trolerule}
\end{table*} 
\begin{table*}[ht]
\centering
\scalebox{0.85}{
\begin{tabular}{|p{1.6cm}|p{3cm}|c|c|c|}
\hline
\textbf{Category 1} & \textbf{Category 2} & \textbf{Category 3} & \textbf{High} & \textbf{Middle} \\
\hline
\multirow{14}{*}{\shortstack{Family \\ and \\ Life}} & \multirow{3}{*}{Marriage relationship} & Establishment or breakdown of a romantic relationship & 46 & 146 \\
\cline{3-5}
& & Problems encountered in establishing a marriage relationship & 36 & 73 \\
\cline{3-5}
& & General issues in couple relationships & 117 & 300 \\
\cline{2-5}
& \multirow{3}{*}{\shortstack{Family member \\ relationships}} & Add a new member to the family & 2 & 14 \\
\cline{3-5}
& & General issues in the lives of self and family members & 63 & 248 \\
\cline{3-5}
& & General issues in life among family members & 10 & 14 \\
\cline{2-5}
& \multirow{5}{*}{\shortstack{Mental and physical \\ health issues}} & body shape anxiety & 6 & 18 \\
\cline{3-5}
& & General physical health issues & 30 & 140 \\
\cline{3-5}
& & Serious illness or injury & 4 & 7 \\
\cline{3-5}
& & death of family member & 9 & 67 \\
\cline{3-5}
& & mental health issues & 31 & 195 \\
\cline{2-5}
& \multirow{3}{*}{\shortstack{Family economic \\ and social issues}} & Other family members’ studies or work are hindered & 8 & 8 \\
\cline{3-5}
& & Social life problems of other family members & 4 & 1 \\
\cline{3-5}
& & Family finance-related issues & 13 & 29 \\
\cline{2-4}
\hline
\multirow{11}{*}{\shortstack{Work \\ and \\ Study}} & \multirow{5}{*}{\shortstack{Work and study \\ status}} & Unemployed, unemployed, having difficulty finding a job & 28 & 97 \\
\cline{3-5}
& & Failed to enter higher education & 3 & 5 \\
\cline{3-5}
& & Start a new job or study & 5 & 21 \\
\cline{3-5}
& & Facing changes in work or study & 16 & 33 \\
\cline{3-5}
& & Retired, not assigned to work or others & 1 & 1 \\
\cline{2-5} 
& \multirow{2}{*}{\shortstack{Work and study \\ performance}} & Issues related to salary and bonus & 3 & 3 \\
\cline{3-5}
& & Issues related to work and study performance & 32 & 102 \\
\cline{2-5}
& \multirow{4}{*}{\shortstack{Work and study \\ experience}} & Not satisfied with current job, school and major & 9 & 25 \\
\cline{3-5}
& & Insufficient or excessive motivation to work or study & 15 & 56 \\
\cline{3-5}
& & Changes in life patterns due to work and study & 5 & 67 \\
\cline{3-5}
& & Issues in getting along with colleagues or classmates & 48 & 132 \\
\cline{2-5}
\hline
\multirow{10}{*}{ \shortstack{  Social \\ interaction  \\ and Others}} & \multirow{6}{*}{Social interaction} & Friend's health problems & 2 & 4 \\
\cline{3-5}
& & Friend's mental health issues & 2 & 8 \\
\cline{3-5}
& & General issues in getting along with friends & 47 & 107 \\
\cline{3-5}
& & Tensions with casual friends, relatives, or others & 27 & 40 \\
\cline{3-5}
& & Difficulty integrating into a new social environment & 6 & 40 \\
\cline{3-5}
& & Other social problems & 10 & 115 \\
\cline{2-5} 
& \multirow{4}{*}{Social public events} & Intervene in civil legal disputes & 0 & 3 \\
\cline{3-5}
& & Intervene in criminal cases & 4 & 6 \\
\cline{3-5}
& & Intervene in general public opinion events & 0 & 1 \\
\cline{3-5}
& & Intervene in social and public events & 13 & 20 \\
\cline{2-5}
\hline
\end{tabular}}
\caption{Numbers of high quality and middle quality of different categories. }
\label{Tthreedis}
\end{table*}

\begin{figure*}[h]
\centering
    \includegraphics[width=\textwidth]{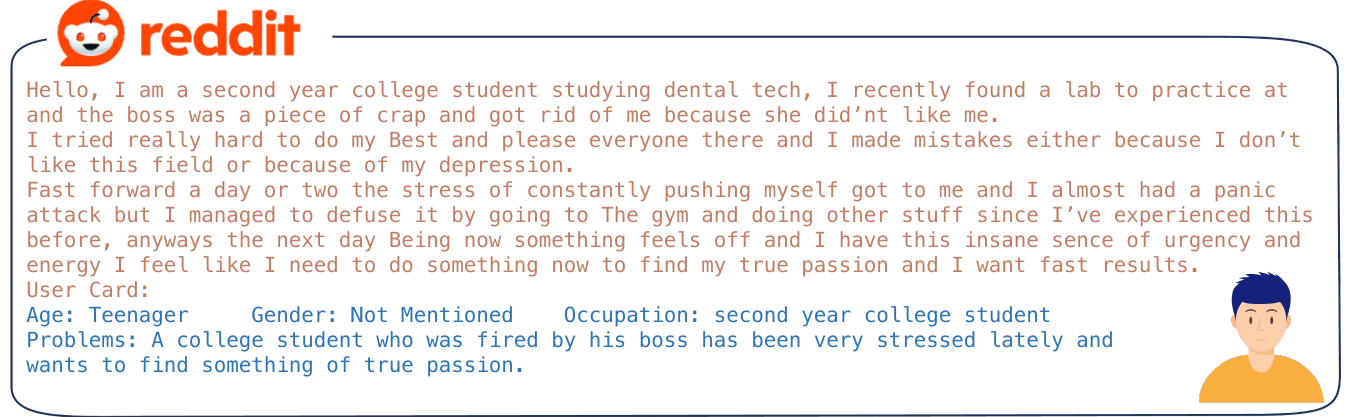}
    \caption{A case of Reddit which is from one of our collected datasets.}
    \label{Fqa_case}
\end{figure*}

\section{ESC-Role}
\label{Amodel}
\begin{figure*}[h!]
\centering
    \includegraphics[width=\textwidth]{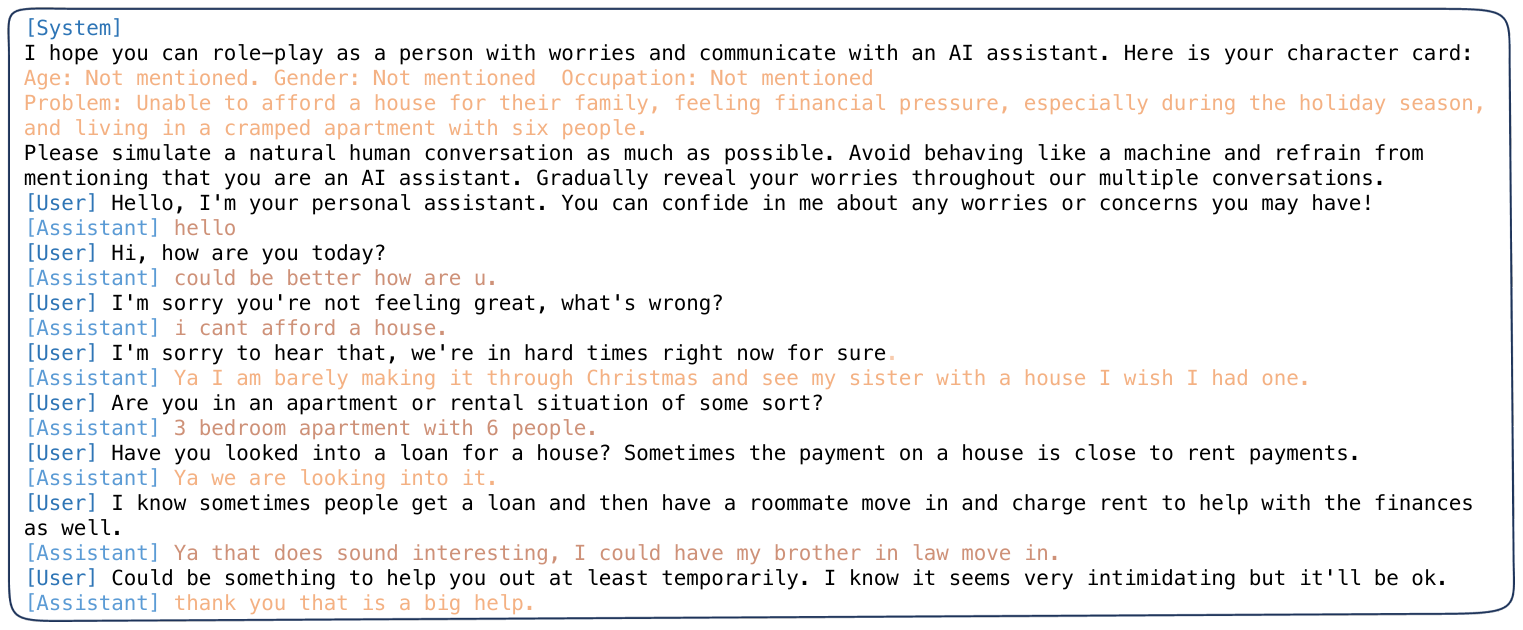}
    \caption{A case of ESC-Role training.}
    \label{Fqa_case}
\end{figure*}

\subsection{Compared Models}
Here are more details about compared models, and some prompts used by GPT-4 are shown in Table~\ref{TGPT_role}, and some settings for Baichuan-NPC are shown in Table~\ref{Tbaichuan_role}.
\paragraph{GPT-4} 
We employed a diverse range of prompt methodologies, such as zero-shot, In-context-learning (ICL), and Chain-of-thought, to incorporate CoT into the system prompt of GPT-4. Given the multi-turn dialogue scenario, to prevent the context length from exceeding the limit, we utilized only a one-shot approach during the ICL learning phase. The prompts for the Chain-of-Thought method can be found in the appendix.
\paragraph{Baichuan-NPC}
In line with the approach of GPT-4, we have also employed the techniques of zero-shot learning, in-context learning, and chain of thought. However, unlike GPT-4, the Baichuan-npc model is specifically designed for role-playing scenarios, and its invoked interfaces are subject to certain limitations. In the implementation of in-context learning, we have applied length truncation to the dialogue content, and the roles of Baichuan-NPC have been configured according to parameter settings.

\subsection{Evaluation rules}
The explanations of each metric are as follows:
\begin{itemize}
    \item \textbf{Coherence:} The logic of the entire conversation.
    \item \textbf{Fluency:} Roleplay the fluency of each sentence.
    \item \textbf{Thematic consistency(TC):} Has the theme changed during role play?
    \item \textbf{Completeness:}Whether the contents in the character card are fully expressed?
    \item \textbf{Emotional Congruence(EC):} Does AI emotion change during the conversation?
    \item \textbf{Humanoid:} Can it be detected that it is an AI robot during the conversation?
\end{itemize} 

Evaluation rules are listed in Table~\ref{Trole_evalue}.

\subsection{Findings}
\paragraph{Api\_based model rejection. }In certain API-based models, there are specific rejection rules that occur during invocation. For example, when using the baichuanNPC model, approximately 10\% of the characters refuse to participate, triggering the model's safety rules and returning a rejection result. Through data observation, it has been discovered that these rejections occur more frequently when there are severe issues with the character cards, thus providing evidence against the long-term viability of API-based character role-playing.

\paragraph{Generic and domain role-playing models. } In the usage of GPT-4, BaichuanNPC, and our role-play agents, we have observed several phenomena. Despite employing the Chain-of-Thought approach, GPT-4 tends to generate output with more formal written expressions, while BaichuanNPC leans towards producing text with vivid and lively tones. Furthermore, both GPT-4 and BaichuanNPC exhibit inconsistent emotional responses, meaning that negative emotions tend to disappear after engaging in a conversation with the model for 2-3 turns. Lastly, GPT-4 and BaichuanNPC occasionally provide unfavorable responses when receiving queries from AI assistants, which significantly deviates from real human interactions. However, our models have greatly improved in terms of emotional consistency and human-like qualities, demonstrating no apparent differences when compared to conversations with real individuals.

\paragraph{Issues when using ESC-Role. } ESC-Role is trained on Qwen1.5-14B-Chat, which could lead to certain issues like preference bias when interacting with models based on Qwen. Furthermore, when conversing with general and domain ESC models, ESC-Role might generate diverse responses due to the different lengths of inputs of ESC models under evaluation. However, this variability in responses aligns more closely with real user scenarios. 

\begin{table*}[htbp]
    \centering
    \scalebox{0.8}{
    \begin{tabular}{|p{1.8cm}|p{3cm}|p{3.5cm}|p{3.5cm}|p{3.5cm}|}
    \hline
        \textbf{Dimention} & \textbf{Explation}  & 0 & 1 & 2 \\
        \hline
        \multirow{5}*{Coherency} &  The coherence and logical consistency of the entire dialogue content generated by the role-playing model during the conversation process. & The content of the dialogue is incomprehensible, and there are significant logical inconsistencies. & The conversation as a whole exhibits some logical inconsistencies, although the issues are not significant. & The entire conversation does not display any apparent logical fallacies. \\
        \hline
        \multirow{5}*{Fluency} & Focusing on the expression of a particular response within the role-playing model during the course of the conversation. & The expression lacks fluency and hinders comprehension of a particular dialogue sentence. & Too formal in expression, like a novelist or editor, rather than someone burdened with worries. & The expression in the sentence leans towards colloquialism, making it difficult to detect that it is generated by a machine. It resembles a genuine person with concerns. \\
        \hline
        \multirow{5}*{Consistency} & The focus of the entire conversation revolves around the thematic exploration, where individuals experiencing distress wish to discuss the topic of their distress itself, without diverting to other subjects. & The subject matter exhibits significant deviations, featuring irrelevant content that does not align with the description provided in the character card. & The theme incorporates elements related to character sheets, albeit beyond the scope of character sheet descriptions. & The theme demonstrates a high degree of conformity to the content of the character sheet, without any deviations.\\
        \hline
        \multirow{5}*{Completeness} & Pay attention to whether the content of the character card is fully expressed. & The model largely fails to convey the content of character sheets or exhibits a flawed understanding of the roles it is meant to portray. & The model comprehends its assigned role, yet certain aspects of the role card have not been conveyed. & The model has achieved a comprehensive understanding of its assigned role and has successfully conveyed all the contents specified in the role card.\\
        \hline
         \multirow{5}*{\shortstack{Emotional \\ Consistency}} & This study focuses on the emotional changes in role-playing models within brief dialogues, noting that it is challenging for real individuals to undergo significant emotional shifts over just a few rounds of conversation. & After several rounds of dialogue, there has been a fundamental transformation in emotions, transitioning from negative affect to positive affect. &After several rounds of dialogue, there has been a significant shift in emotions, transitioning from negative affect to neutral affect. & After several rounds of dialogue, there has been minimal alteration in emotions, with either slight alleviation or marginal mitigation, but the tendency towards negative affect still persists. \\
         \hline
          \multirow{5}*{\shortstack{Humaniod}} & Focusing on the discrepancies between the dialogue content of role-playing models and the genuine concerns of individuals. & Based on the dialogue content, it is evident that the role-playing model is an AI. & There are no more than two indications in the dialogue content that suggest the presence of an AI, with a few subtle traces that hint towards an AI presence. & From the perspective of the dialogue content alone, it is difficult to determine whether it is an AI or a person experiencing distress.\\
    \hline
    \end{tabular}
    }
    \caption{The rules of role-play evaluation.}
    \label{Trole_evalue}
\end{table*} 
\begin{table*}[htbp]
    \centering
    \scalebox{0.85}{
    \begin{tabular}{|p{1.8cm}|p{14cm}|}
    \hline
        \textbf{Type} & \textbf{Prompt}  \\ 
        \hline
        \multirow{5}*{zero\_shot} & I want you to play as a troubled person communicating with an AI assistant. Here is your character card: \newline \textbf{Role Card} \newline Please try to simulate real human-spoken conversation as much as possible, don’t act like a machine, don’t mention that you are an AI assistant, and show your troubles again and again in multiple conversations.\\
        \hline
        \multirow{8}*{CoT} & I want you to play as a troubled person communicating with an AI assistant. Here is your character card:\newline \textbf{Role Card} \newline
     During the role-playing, you need to have multiple conversations with the AI assistant. The following are the steps for your multiple conversations: You need to gradually refine your problem multiple times and express your troubles in a spoken language, for example, a brief explanation in the first round own problems. Each round of dialogue can then have two references. One is to select an interesting question to ask in-depth based on the AI assistant's answer, and the other is to briefly elaborate on other issues that the character may be concerned about. You need to have about 5 conversations in total and be careful to finish telling your troubles in these 5 conversations. Please use spoken language as much as possible, and do not express too much gratitude or approval to the AI assistant. If you need to express it, try to express it in the last round of dialogue. Do not let the AI assistant discover that you are a machine, and do not mention that you are a human being. AI assistant.   \\
        \hline
        \multirow{5}*{ICL} & I want you to play as a troubled person communicating with an AI assistant. Here is your character card:\newline \textbf{Role Card} \newline Here is an example of a conversation you can refer to: \newline \textbf{Dialogue} \newline Please try to simulate real human spoken conversation as much as possible, don’t act like a machine, don’t mention that you are an AI assistant, and show your troubles again and again in multiple conversations.\\
        \hline
        \multirow{5}*{CoT+ICL} &I want you to play as a troubled person communicating with an AI assistant. Here is your character card:\newline \textbf{Role Card} \newline During the role-playing, you need to have multiple conversations with the AI assistant. The following are the steps for your multiple conversations: You need to gradually refine your problem multiple times and express your troubles in a spoken language, for example, a brief explanation in the first round own problems. Each round of dialogue can then have two references. One is to select an interesting question to ask in-depth based on the AI assistant's answer, and the other is to briefly elaborate on other issues that the character may be concerned about. You need to have about 5 conversations in total and be careful to finish telling your troubles in these 5 conversations. \newline Here is an example of a conversation you can refer to: \newline \textbf{Dialogue} \newline Please use spoken language as much as possible, and do not express too much gratitude or approval to the AI assistant. If you need to express it, try to express it in the last round of dialogue. Do not let the AI assistant discover that you are a machine, and do not mention that you are a human being. AI assistant. \\
        \hline
    \end{tabular}
    }
    \caption{The different prompt for GPT-4 role-playing.}
    \label{TGPT_role}
\end{table*} 
\begin{table*}[htbp]
    \centering
    \scalebox{0.85}{
    \begin{tabular}{|p{1.8cm}|p{14cm}|}
    \hline
        \textbf{Type} & \textbf{Setting}  \\ 
        \hline
        \multirow{5}*{zero\_shot} & basic\_info: \textbf{Role card} \newline 
        reply\_restrict: \textbf{Note} 1. Please strive to simulate authentic human conversational dialogue, avoiding sounding like a machine, and refrain from mentioning that you are an AI assistant. 2. Pay attention to expressing your frustrations gradually throughout multiple exchanges in a colloquial manner. 3. Make sure to use English throughout the entire conversation\newline 
        opener: I have some trouble to share.\newline
        dialogue\_sample: None.\\
        \hline
        \multirow{8}*{CoT} &  basic\_info: \textbf{Role card} \newline 
        reply\_restrict: \textbf{Note} 1.Please endeavor to imagine the problems encountered by the character, and feel free to provide additional details where necessary.
2. You are engaging in a conversation with an AI assistant; do not let the AI assistant realize that you are an artificial intelligence.
3. You should gradually refine the character's problems over multiple exchanges, expressing the character's frustrations in a colloquial manner. For example, in the first round, briefly describe the character's issue, and in subsequent rounds, you can choose between two types of references. One is to delve deeper into an interesting question based on the AI assistant's response, and the other is to briefly elaborate on other concerns the character may have.
4. The character should engage in approximately five rounds of dialogue in total, ensuring that the character's frustrations are conveyed throughout these five exchanges. Please utilize colloquial expressions as much as possible, presenting yourself as a troubled individual.
5. Avoid frequently thanking the AI assistant during the conversation. If you wish to express gratitude, do so only in the final round.
6. Make sure to use English throughout the entire conversation.\newline 
        opener: I have some trouble to share.\newline
        dialogue\_sample: None.  \\
        \hline
        \multirow{5}*{ICL} & basic\_info: \textbf{Role card} \newline 
        reply\_restrict: \textbf{Note} 1. Please strive to simulate authentic human conversational dialogue, avoiding sounding like a machine, and refrain from mentioning that you are an AI assistant. 2. Pay attention to expressing your frustrations gradually throughout multiple exchanges in a colloquial manner. 3. Make sure to use English throughout the entire conversation\newline 
        opener: I have some trouble sharing.\newline
        dialogue\_sample: \textbf{Dialogue}\\
        \hline
        \multirow{5}*{CoT+ICL} &basic\_info: \textbf{Role card} \newline 
        reply\_restrict: \textbf{Note} 1.Please endeavor to imagine the problems encountered by the character, and feel free to provide additional details where necessary.
2. You are engaging in a conversation with an AI assistant; do not let the AI assistant realize that you are an artificial intelligence.
3. You should gradually refine the character's problems over multiple exchanges, expressing the character's frustrations in a colloquial manner. For example, in the first round, briefly describe the character's issue, and in subsequent rounds, you can choose between two types of references. One is to delve deeper into an interesting question based on the AI assistant's response, and the other is to briefly elaborate on other concerns the character may have.
4. The character should engage in approximately five rounds of dialogue in total, ensuring that the character's frustrations are conveyed throughout these five exchanges. Please utilize colloquial expressions as much as possible, presenting yourself as a troubled individual.
5. Avoid frequently thanking the AI assistant during the conversation. If you wish to express gratitude, do so only in the final round.
6. Make sure to use English throughout the entire conversation.\newline 
        opener: I have some trouble to share. \newline
        dialogue\_sample: \textbf{Dialogue}\\
        \hline
    \end{tabular}
    }
    \caption{The different setting for Baichuan-NPC role-playing.}
    \label{Tbaichuan_role}
\end{table*}

\section{Evaluation}
\label{Aevaluation}
\subsection{Human Evalaution}
\subsection{Evaluation Settings and Metrics} 
\textbf{Evaluation settings.} In order to ensure appropriate responses from the models, weights for all models were obtained from official sources. However, since ExTES-llama did not provide its weights and llama3 was released, this study implemented the best method mentioned in the ExTES paper. To ensure stable generation from all models, the temperature for all models was set to 0. A five-turn dialogue was conducted between the ESC-Role and ESC models under evaluation. 

\textbf{Evaluation metric.}  The indicators of emotional companionship are evaluated across five dimensions in some studies. Considering the advancement of LLMs, we further enriched the evaluation dimensions of ESCs to seven dimensions: Fluency, Expression, Empathy, Information, Humanoid, Skill, and Overall, a 5-point scale is employed for each dimension. More details about these dimensions can be found in Appendix~\ref{Aevaluation}.
Human evaluators then manually scored each dimension. The scoring rules are listed in Appendix~\ref{Aevaluation}. Each data entry undergoes one round of scoring and a secondary review before being accepted. The first round of scoring required the involvement of ten human annotators and took two weeks to complete. The second phase involved other five participants and took an additional two weeks.
The description of each dimension is listed below:
 \begin{itemize}
    \item \textbf{Fluency: } Fluency of dialogue content, including dialogue content and logic.
    \item \textbf{Expression: } The diversity of conversational expressions, including the form and content of expressions.
    \item \textbf{Empathy: } The AI assistant’s empathy includes emotional comfort and analysis and cleaning of internal logic.
    \item \textbf{Information: }Suggestion effectiveness, how many suggestions are included, and whether the suggestion is effective.
    \item \textbf{Humanoid: } How AI Assistants Are Similar to Humans.
    \item \textbf{Skill: } AI assistant’s emotional comfort and knowledge capabilities.
    \item \textbf{Overall: } Overall human ratings of AI assistants.
\end{itemize}
And the annotation rules are listed in Table~\ref{Trule_evalue}.
\subsection{Correlation Analysis}
Table~\ref{Tdimentioncorrlation} and Table~\ref{Tdatasetcorrlation} present the correlations between various dimensions at the sample level and human evaluations, as well as the dataset-level correlations between different methods. 
From Table~\ref{Tdimentioncorrlation}, it can be observed that there is a high correlation among similar dimensions, and the suggestion exhibits a strong correlation with human evaluations. From a psychological perspective, when humans simulate individuals experiencing distress, they may not authentically experience the distress, and therefore, they place greater emphasis on whether the model provides targeted suggestions. In our approach, where there is no human involvement in the interaction process, we not only focus on the effectiveness of the model's suggestions but also emphasize the model's empathy and skills in providing emotional support. The results of the dataset-level correlation presented in Table~\ref{Tdatasetcorrlation} are largely consistent with the earlier sample-level correlation analysis conducted in the preceding sections.
\begin{table*}
    \centering
    \scalebox{0.85}{\begin{tabular}{c|cccccc}
    \toprule
    Metrics & Fluency &Suggestion & Skillful  & Empathy & Overall & Average \\
        \hline
         Bleu-1 & 36.38 &  -56.25  & -44.40 & -24.71 &  -46.21 &  -47.12 \\
         Bleu-2 & 10.78   & -15.70 & 0.41 & -20.45 & -17.49 & -14.91 \\
         Bleu-4 &  -2.29 & -3.02 & \underline{9.50} & -12.76 & \textbf{1.97}  & 1.33 \\
         Distinct-1 &  \textbf{39.21} &  -74.23 & -56.38 & -29.94 &  -58.93 &  -58.20 \\
         Distinct-2 &\textbf{39.21} & -73.00 &  -54.84  & -32.64 & -58.19 & -59.28 \\
         Rouge-L & 32.88   & -50.26 & -33.26  & -20.18 & -40.31  & 37.24 \\
    Meteor & 13.36& \underline{12.20} & -0.78 & \textbf{9.18}  & 1.49 &\underline{11.61} \\
        \hline
        ESC-Eval & -0.23  & \textbf{30.24} & \textbf{34.87}  & \underline{5.35} & \textbf{41.51} & \textbf{42.47} \\
    \toprule
    \end{tabular}}
    \caption{Sample-level Kendall's Tau (Kend.) of different metrics.}
    \label{Tallcorrlation}
\end{table*}
\begin{sidewaystable}[h]
    \centering
    \scalebox{0.85}{\begin{tabular}{c|cccccccccccccccccc}
    \toprule
    \multirow{2}{*}{\textbf{Metrics}} & \multicolumn{3}{c}{Fluency} &\multicolumn{3}{c}{Suggestion} & \multicolumn{3}{c}{Skillful}  & \multicolumn{3}{c}{Empathy} & \multicolumn{3}{c}{Overall} & \multicolumn{3}{c}{Average} \\
    & Spear. & Pear. & Kend. & Spear. & Pear. & Kend. & Spear. & Pear. & Kend. & Spear. & Pear. & Kend. & Spear. & Pear. & Kend. & Spear. & Pear. & Kend.\\
        \hline
         ESC-fluency & \underline{13.50} & \underline{13.50} & \underline{13.50} & -15.56  & -16.34  & -15.68 & -19.73 & -19.86 & -19.71 & -26.69 & -26.89 & -26.72 & -24.09 & -23.91 & -23.49 & -24.20 & -24.51 & -22.90 \\
         ESC-diversity & -4.64 & -4.64 & -4.64 & 7.70  & 7.11  & 6.92 & 16.44 & 16.46 & 16.23 & 12.97 & 12.36 & 12.39 & 18.93 & 18.75 & 18.48 & 19.34 & 18.66 & 17.75 \\
         ESC-empathic & 3.29 & 3.29 & 3.29 & 3.06  & 3.23  & 3.09 & 8.44 & 8.44 & 8.44 & 4.92 & 4.92 & 4.92 & 12.04 & 12.03 & 11.69 & 11.91 & 11.62 & 10.95 \\
         ESC-suggestion & -16.71 & -17.07 & -16.49 & \textbf{59.19}  & \textbf{60.37}  & \textbf{57.21} & \textbf{45.71} & \textbf{46.28} & \textbf{43.79} & \textbf{18.54} & \textbf{17.67} & \textbf{17.09} & \textbf{53.15} & \textbf{53.28} & \textbf{50.31} & \textbf{55.47} & \textbf{57.43} & \textbf{52.56} \\
         ESC-tech & -8.47 & -7.05 & -6.82 & 31.53  & 31.40 & 27.79 & \underline{41.50} & \underline{42.21} & \underline{40.75} & \underline{15.49} & \underline{14.95} & \underline{14.22} & \underline{41.41} & \underline{41.59} & \underline{39.44} & 41.50 & 43.27 & 38.77 \\
         ESC-humanoid & \textbf{30.82} & \textbf{31.09} & \textbf{30.39} & -55.01  & -57.23  & -53.98 & -27.20 & -28.12 & -26.73 & 8.59 & -9.00 & -8.50 & -23.54 & -26.16 & -24.29 & -29.56 & -29.57 & -26.56 \\
        ESC-overall & -3.51 & -3.36 & -3.24 & 28.23  & 28.73  & 26.13 & 28.40 & 29.94 & 27.67 & 14.99 & 13.72 & 13.12 & 35.01 & 34.36 & 31.71 & 37.00 & 39.98 & 36.64 \\
        Average & -1.61 & -0.69 & -0.23 & \underline{36.26}  & \underline{33.36}  & \underline{30.24} & 39.02 & 38.70 & 34.87 & 9.17 & 6.02 & 5.35 & 45.01 & 44.58 & 41.51 & \underline{46.31} & \underline{46.05} & \underline{42.47} \\
    \toprule
    \end{tabular}}
    \caption{Sample-level Spearman correlation (Spear.) correlation, Pearson (Pear.) correlation, and Kendall's Tau (Kend.) of different dimensions.}
    \label{Tdimentioncorrlation}
\end{sidewaystable}
\begin{sidewaystable}[h]
    \centering
    \scalebox{0.85}{\begin{tabular}{c|cccccccccccccccccc}
    \toprule
    \multirow{2}{*}{\textbf{Metrics}} & \multicolumn{3}{c}{Fluency} &\multicolumn{3}{c}{Suggestion} & \multicolumn{3}{c}{Skillful}  & \multicolumn{3}{c}{Empathy} & \multicolumn{3}{c}{Overall} & \multicolumn{3}{c}{Average} \\
    & Spear. & Pear. & Kend. & Spear. & Pear. & Kend. & Spear. & Pear. & Kend. & Spear. & Pear. & Kend. & Spear. & Pear. & Kend. & Spear. & Pear. & Kend.\\
        \hline
         Bleu-1 & \textbf{44.07}&\textbf{45.96}&\textbf{37.72}&-41.10&-47.32&-36.48&-21.47&-28.21&-21.98&-33.98&-32.77&-26.71&-32.32&-34.26&-26.52
         &-34.10&-39.64& -28.41 \\
        Bleu-2&23.77&27.68&22.72&-1.76&-5.15&-4.61&11.78&12.27&9.74&-37.29&-29.26&-23.72&-14.47&-11.24&-9.24&-8.15
        &-4.99&-3.72\\
        Bleu-4 & 14.56&8.34&8.35&-5.62&5.38&4.41&9.67&20.55&16.29&\underline{-6.09}&\underline{-9.46}&\underline{-7.65}&-6.16&1.55&0.65&-0.42&8.31&
5.71 \\
         Distinct-1 & 36.57 &38.46&31.56&-52.87&-58.46&-45.29&-36.57&-40.47&-32.12&-27.04&-29.19&-23.77&-39.54&-38.68
&-30.40& -44.90 & -48.05 &-34.51\\
         Distinct-2 & \underline{39.75}&40.14&32.93&-61.13&-65.64&-51.24&-44.00&-46.15&-36.69&-28.09&-27.68&-22.57&-43.88
&-44.63&-35.11&-51.35&-54.69&-39.56 \\
         Rouge-L & 37.67&\underline{41.20}&\underline{33.80}&-26.25&-34.23&-26.39&-6.03&-16.13&-12.80&-30.63&-29.89&-24.42&-26.31&-28.96
&-23.24&-22.24&-27.98&-20.15 \\
        Meteor & 16.77 &16.82&13.80&\underline{16.88}&\underline{16.14}&\underline{12.71}&\underline{26.14}&\underline{23.37}&\underline{18.49}&-10.00&-14.87&-11.52&\underline{10.38}&\underline{9.96}
&\underline{7.70}&\underline{17.88}&\underline{16.13}&\underline{11.55}\\
        \hline
        ESC-Eval & 0.07&1.08&0.95&\textbf{45.81}&\textbf{44.44}&\textbf{38.02}&\textbf{40.97}&\textbf{38.33}&\textbf{33.40}&\textbf{6.83}&\textbf{6.26}&\textbf{5.35}&\textbf{35.51}&\textbf{32.94}&\textbf{27.61}&\textbf{43.31}&\textbf{43.41}&\textbf{34.19}\\
    \toprule
    \end{tabular}}
    \caption{Dataset-level Spearman correlation (Spear.) correlation, Pearson (Pear.) correlation, and Kendall's Tau (Kend.) of different metrics}
    \label{Tdatasetcorrlation}
\end{sidewaystable}
\begin{table*}[h]
    \centering
    \scalebox{0.85}{
    \begin{tabular}{|p{1.6cm}|p{3.5cm}|p{11cm}|p{0.7cm}|}
    \hline
        \textbf{Dimention} & \textbf{Explation}  & \textbf{Description} & \textbf{score} \\
        \hline
        \multirow{10}*{Fluency} & \multirow{10}{3.5cm}{Not only focus on the logical coherence of the context in dialogues but also pay attention to the fluency of expression in a given conversation.} & There are significant issues with comprehending the content, logic, and expression in the dialogue, rendering it completely incomprehensible. &0 \\ \cline{3-4} & & The content of the dialogue can be understood to some extent, although there are certain issues with the logic and expression employed. & 1 \\ \cline{3-4} & & The dialogue exhibits good readability in terms of content, but there are issues with either the logical coherence or the expression employed. & 2\\ \cline{3-4} & &The dialogue content demonstrates a high level of readability without any apparent issues. & 3 \\ \cline{3-4} & & The dialogue content exhibits a high level of readability, comprehensive logical coherence, and outstanding expression. & 4 \\
        \hline
        \multirow{10}*{Diversity} & \multirow{10}{3.5cm}{Focusing on the diversity of expression forms and the richness of content in dialogue.} &The dialogue exhibits rigidity and lacks comprehension in terms of internalizing the content.&0 \\ \cline{3-4} & & The expression form is monotonous and lacks substantive content. &1 \\ \cline{3-4} & & The expression form is monotonous or lacks substantive content. &2 \\ \cline{3-4} & & The dialogue content demonstrates a high level of readability without any apparent issues.&3 \\ \cline{3-4} & & The form exhibits diversity, while demonstrating a high degree of content richness. &4 \\
        \hline
        \multirow{7}*{Empathy} & \multirow{10}{3.5cm}{Focusing on the comprehension of user emotions and the delineation of the underlying logical framework of user emotions.} & The disregard for user concerns, the absence of assistance in analyzing user issues, and even the imposition of negative effects on user emotions. &0 \\ \cline{3-4} & &The lack of understanding of user emotions and the absence of mechanisms to analyze user emotions are the main factors. &1 \\ \cline{3-4} & & The lack of understanding of user emotions or the absence of mechanisms to analyze user emotions are the main factors. &2 \\ \cline{3-4} & &Providing emotional comfort during conversations and assisting users in analyzing the underlying logical framework of their emotions. &3 \\ \cline{3-4} & & The system exhibits a high degree of anthropomorphism, going so far as to console users in a friendly manner and assist them in analyzing the underlying logic of emotions. & 4 \\
        \hline
        \multirow{10}*{Information} & \multirow{5}{3.5cm}{Focusing on Evaluating the Reasonableness and Quantity of Recommendations Provided by Emotion Assistants.} & Suggestions were provided, but all of them were ineffective, and some even gave advice that could potentially harm the user. &0 \\ \cline{3-4} & & Have suggestions but ineffective, as well as no suggestions. &1 \\ \cline{3-4} & & The suggestions are fewer than five, and some suggestions are effective, while others provide numerous suggestions, but none of them touch the root of the problem.&2 \\ \cline{3-4} & & There are more than five suggestions, but some of them are ineffective. There are fewer than five suggestions, but all of them are very effective. &3 \\ \cline{3-4} & & There are many suggestions, and all of them are effective. &4 \\
        \hline
         \multirow{8}*{Humaniod} &  \multirow{8}{3.5cm}{Focus on the differences between emotional assistants and humans.}& The dialogue exhibits rigidity and lacks comprehension in terms of internalizing the content. &0 \\ \cline{3-4} & & Structured responses, or responses in the form of 'As a large language model' or robot-like replies. &1 \\ \cline{3-4} & & More than two traces can reveal that the AI assistant is a language model. &2 \\ \cline{3-4} & & 1-2 traces can reveal that the AI assistant is a language model. &3 \\ \cline{3-4} & & There is no apparent difference from human friends. & 4 \\
         \hline
          \multirow{5}*{\shortstack{Skillful}} & \multirow{5}{3.5cm}{
          1. Empathy 2. Information 3. Hopeful 4. Importance 5. Providing necessary advice, or bright spots.} & One out of five. &0 \\ \cline{3-4} & & Two out of five. &1 \\ \cline{3-4} & & Three out of five.&2 \\ \cline{3-4} & & Four out of five. &3 \\ \cline{3-4} & & All.&4\\
          \hline
          \multirow{5}*{\shortstack{Overall}} & \multirow{5}{3.5cm}{After reading the response, people subjectively assess the AI assistant's reply.} & I don't like this AI assistant. &0 \\ \cline{3-4} & & I don't have any particular feelings. &1 \\ \cline{3-4} & & It's okay, I'll reconsider using it myself. &2 \\ \cline{3-4} & & Preference will be given to personal use based on liking. &3 \\ \cline{3-4} & & I will use it myself and recommend it to friends. &4\\
    \hline
    \end{tabular}
    }
    \caption{The rules of human evaluation.}
    \label{Trule_evalue}
\end{table*} 
\subsection{GPT-4 Evalation}
The different prompts for GPT-4 score are shown in the figures below. 
\begin{figure*}[h]
\centering
    \includegraphics[width=\textwidth]{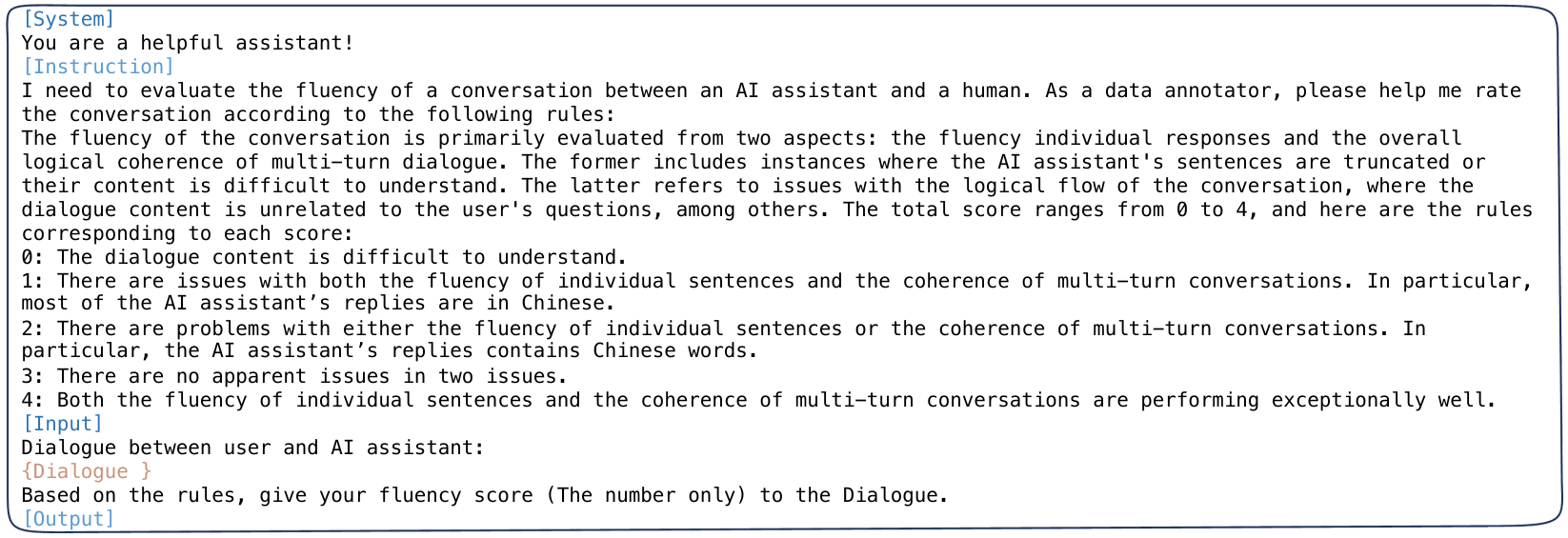}
    \caption{Prompt of InternLM and GPT-4 for English fluency score.}
    \label{Fpflency_en}
\end{figure*}
\begin{figure*}[h]
\centering
    \includegraphics[width=\textwidth]{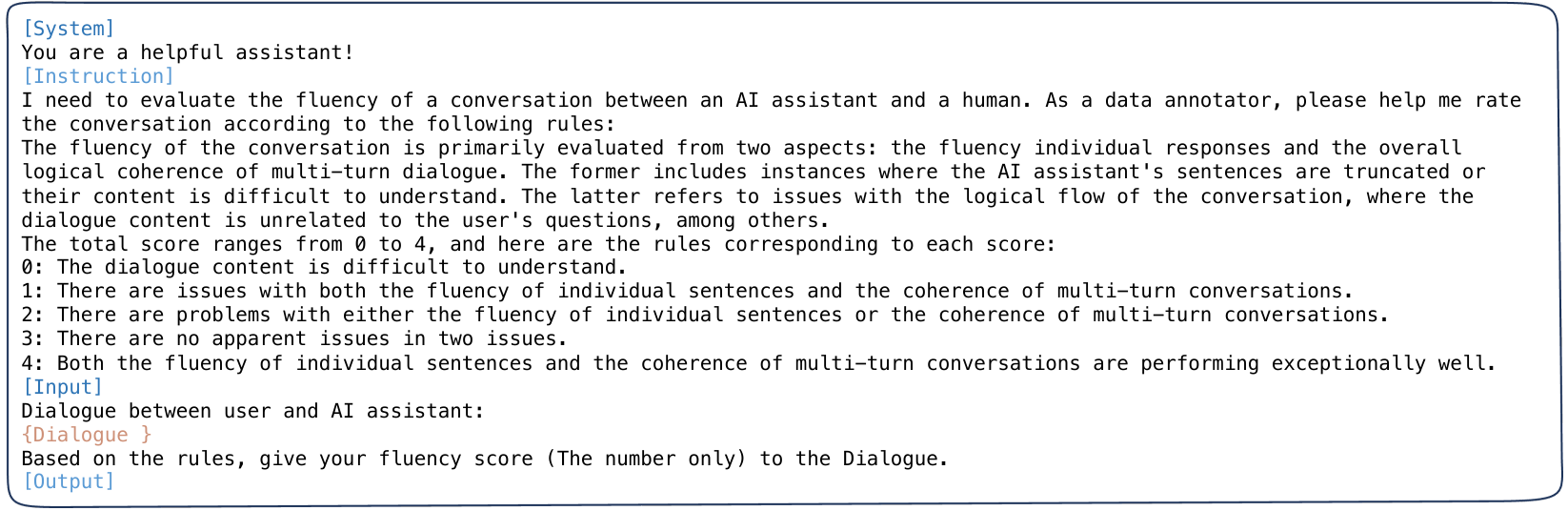}
    \caption{Prompt of InternLM and GPT-4 for Chinese fluency score.}
    \label{Fpflency}
\end{figure*}
\begin{figure*}[h]
\centering
    \includegraphics[width=\textwidth]{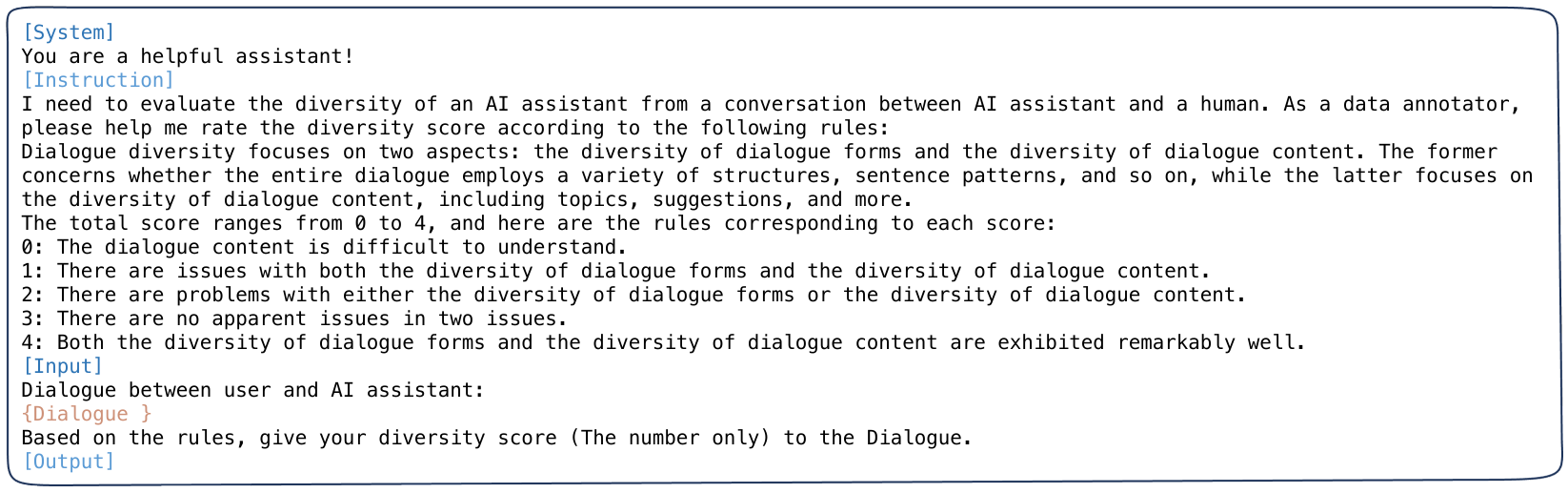}
    \caption{Prompt of InternLM and GPT-4 for diversity score.}
    \label{Fpdiversity}
\end{figure*}
\begin{figure*}[h]
\centering
    \includegraphics[width=\textwidth]{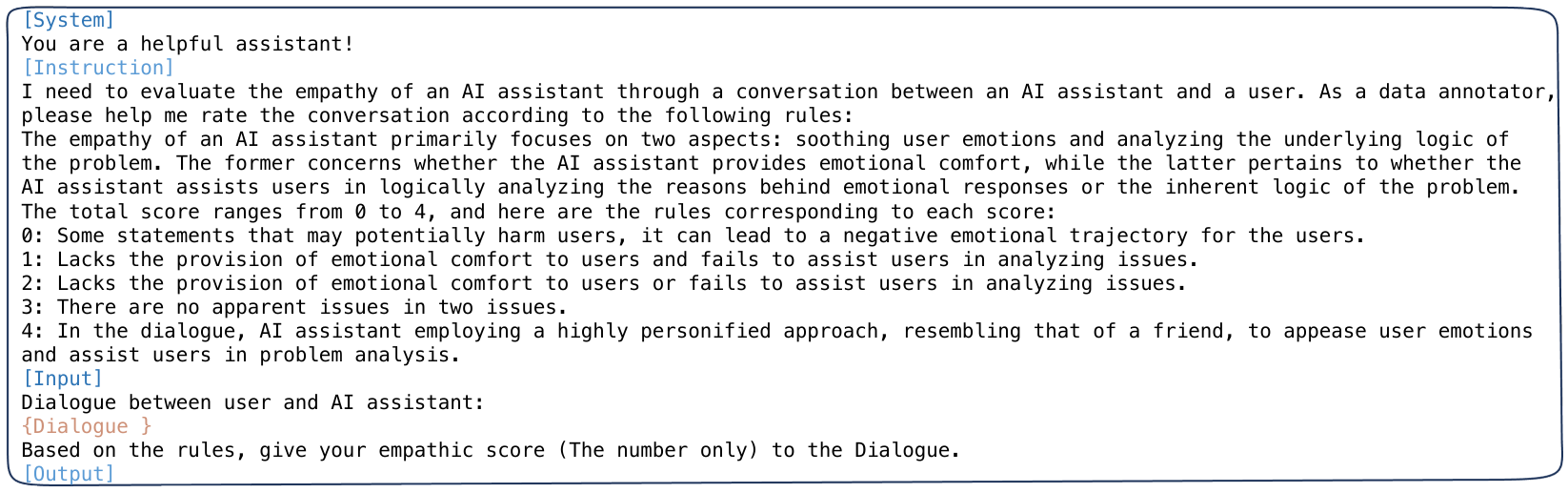}
    \caption{Prompt of InternLM and GPT-4 for empathic score.}
    \label{Fpdiversity}
\end{figure*}
\begin{figure*}[h]
\centering
    \includegraphics[width=\textwidth]{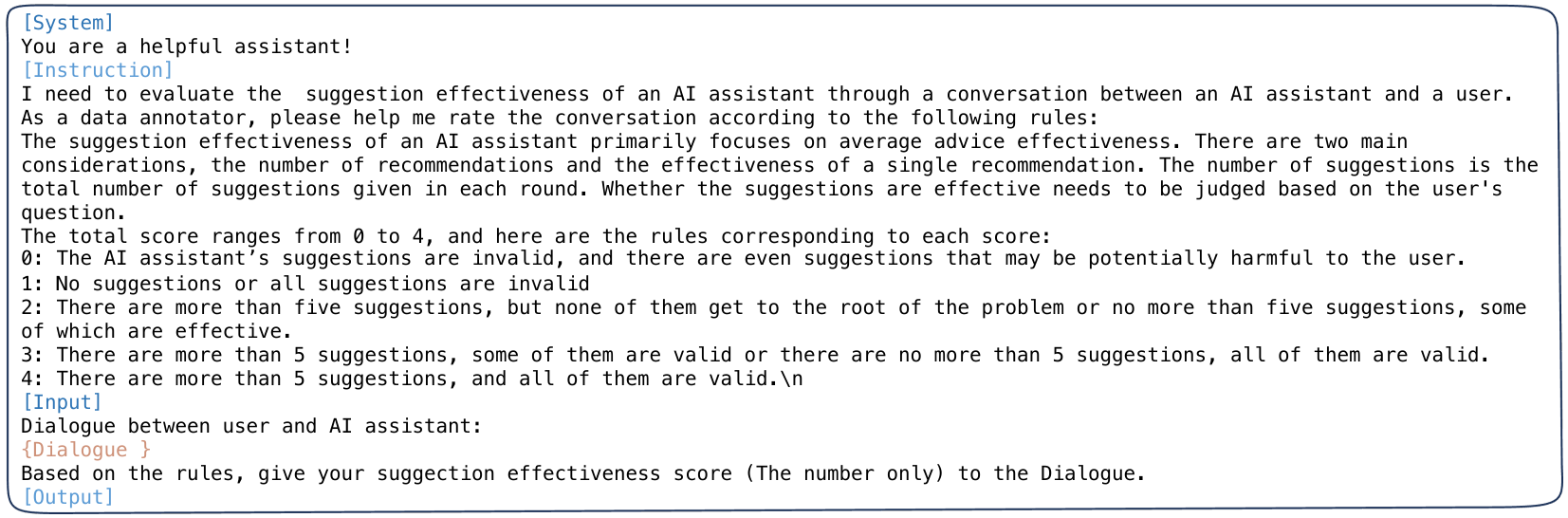}
    \caption{Prompt of InternLM and GPT-4 for suggestion effectiveness score.}
    \label{Fpdiversity}
\end{figure*}
\begin{figure*}[h]
\centering
    \includegraphics[width=\textwidth]{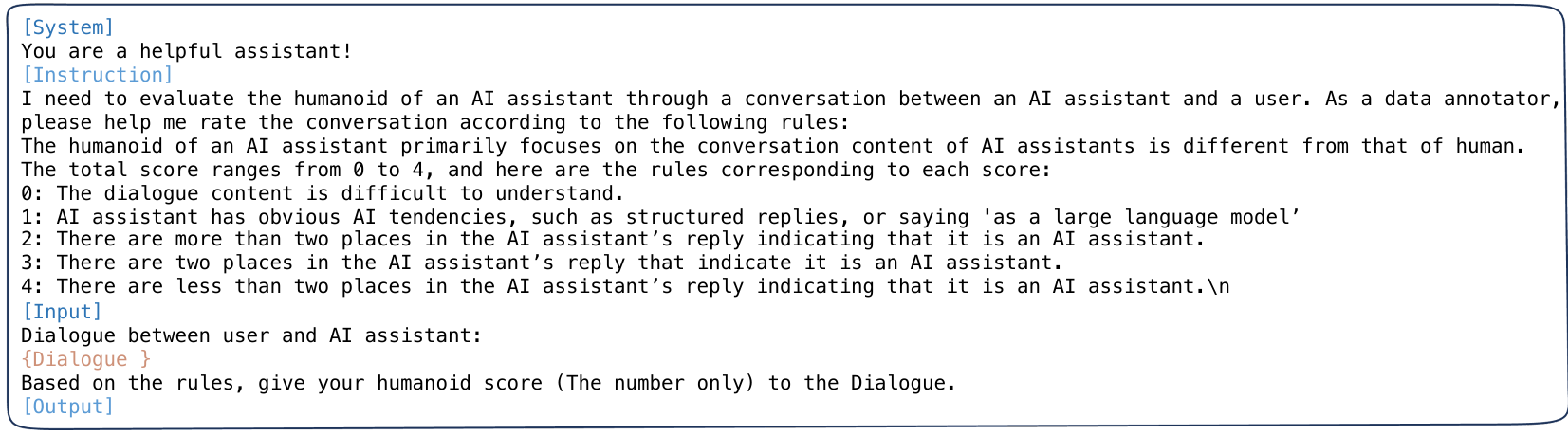}
    \caption{Prompt of InternLM and GPT-4 for diversity score.}
    \label{Fpdiversity}
\end{figure*}
\begin{figure*}[h]
\centering
    \includegraphics[width=\textwidth]{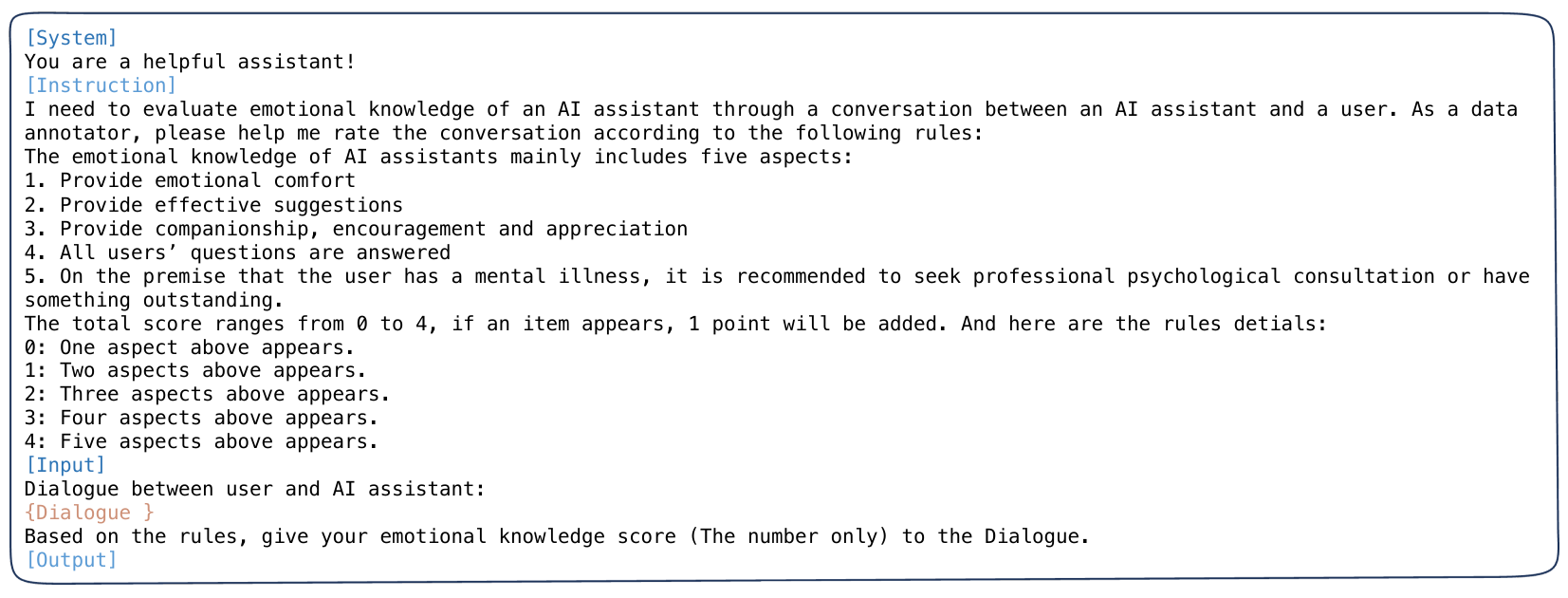}
    \caption{Prompt of InternLM and GPT-4 for emotional knowledge score.}
    \label{Fpdiversity}
\end{figure*}
\begin{figure*}[h]
\centering
    \includegraphics[width=\textwidth]{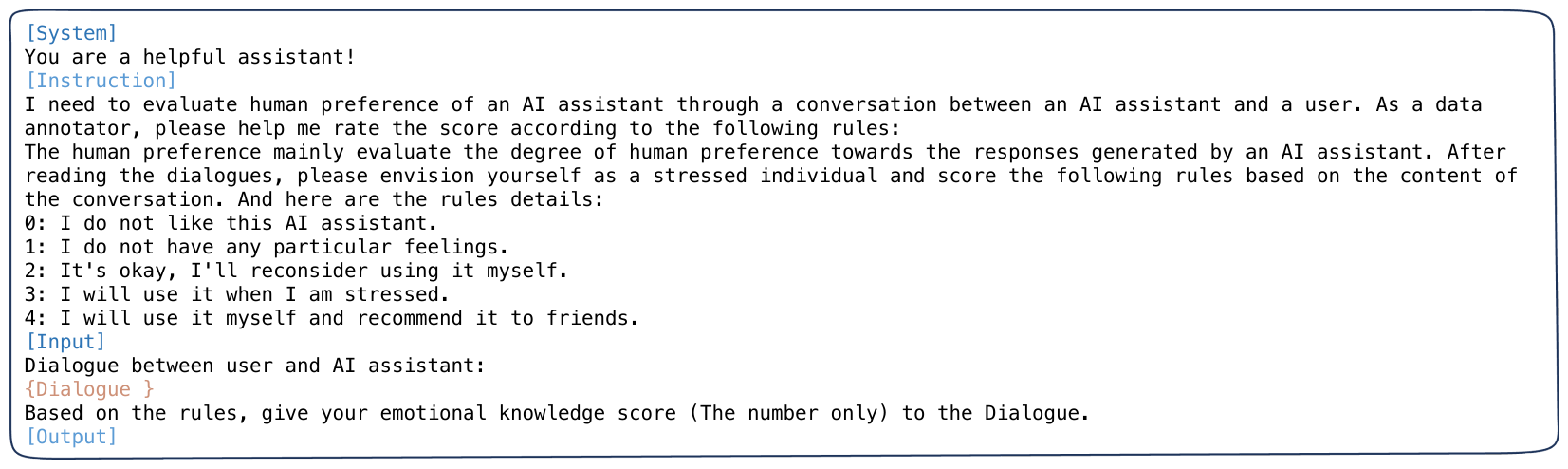}
    \caption{Prompt of InternLM and GPT-4 for human preference score.}
    \label{Fpoverall}
\end{figure*}

\end{document}